\documentclass[sigconf]{acmart}

\copyrightyear{2022}
\acmYear{2022}
\setcopyright{licensedothergov}\acmConference[MM '22]{Proceedings of the
30th ACM International Conference on Multimedia}{October 10--14,
2022}{Lisboa, Portugal}
\acmBooktitle{Proceedings of the 30th ACM International Conference on
Multimedia (MM '22), October 10--14, 2022, Lisboa, Portugal}
\acmPrice{15.00}
\acmDOI{10.1145/3503161.3548363}
\acmISBN{978-1-4503-9203-7/22/10}

\acmSubmissionID{mmfp2821}
\usepackage{graphicx,subfigure,multirow}
\usepackage{soul,xcolor}

\setstcolor{red}

\usepackage{pifont}% http://ctan.org/pkg/pifont
\newcommand{\cmark}{\ding{51}}%
\newcommand{\xmark}{\ding{55}}%

\begin{document}

\title[Bodily Behaviors in Social Interaction: Novel Annotations and State-of-the-Art Evaluation]{Bodily Behaviors in Social Interaction:\\Novel Annotations and State-of-the-Art Evaluation}

\author{Michal Balazia}
\authornote{Authors contributed equally.}
\email{michal.balazia@inria.fr}
\orcid{0000-0001-7153-9984}
\affiliation{%Méditerranée
  \institution{INRIA Sophia Antipolis}
  \streetaddress{2004 Route des Lucioles}
  \city{Sophia Antipolis}
  \country{France}
  \postcode{06902}
}
\author{Philipp Müller}
\authornotemark[1]
\email{philipp.mueller@dfki.de}
\orcid{0000-0001-7037-7100}
\affiliation{%
  \institution{DFKI Saarbrücken}
  \streetaddress{Stuhlsatzenhausweg 3}
  \city{Saarbrücken}
  \country{Germany}
  \postcode{66123}
}
\author{Ákos Levente Tánczos}
\email{akos.tanczos@inria.fr}
\orcid{0000-0002-4760-9599}
\affiliation{%
  \institution{INRIA Sophia Antipolis}
  \streetaddress{2004 Route des Lucioles}
  \city{Sophia Antipolis}
  \country{France}
  \postcode{06902}
}
\author{August von Liechtenstein}
\email{august.liechtenstein@dfki.de}
\orcid{0000-0002-8774-746X}
\affiliation{%
  \institution{DFKI Saarbrücken}
  \streetaddress{Stuhlsatzenhausweg 3}
  \city{Saarbrücken}
  \country{Germany}
  \postcode{66123}
}
\author{François Brémond}
\email{francois.bremond@inria.fr}
\orcid{0000-0003-2988-2142}
\affiliation{%
  \institution{INRIA Sophia Antipolis}
  \streetaddress{2004 Route des Lucioles}
  \city{Sophia Antipolis}
  \country{France}
  \postcode{06902}
}
\renewcommand{\shortauthors}{Michal Balazia et al.} % required

\begin{abstract}
Body language is an eye-catching social signal and its automatic analysis can significantly advance artificial intelligence systems to understand and actively participate in social interactions.
While computer vision has made impressive progress in low-level tasks like head and body pose estimation, the detection of more subtle behaviors such as gesturing, grooming, or fumbling is not well explored.
In this paper we present BBSI, the first set of annotations of complex Bodily Behaviors embedded in continuous Social Interactions in a group setting.
Based on previous work in psychology, we manually annotated 26 hours of spontaneous human behavior in the MPIIGroupInteraction dataset with 15 distinct body language classes.
We present comprehensive descriptive statistics on the resulting dataset as well as results of annotation quality evaluations.
For automatic detection of these behaviors, we adapt the Pyramid Dilated Attention Network~(PDAN), a state-of-the-art approach for human action detection.
We perform experiments using four variants of spatial-temporal features as input to PDAN: Two-Stream Inflated 3D CNN, Temporal Segment Networks, Temporal Shift Module and Swin Transformer.
Results are promising and indicate a great room for improvement in this difficult task.
Representing a key piece in the puzzle towards automatic understanding of social behavior, BBSI is fully available to the research community.

\end{abstract}

%% http://dl.acm.org/ccs.cfm


\begin{teaserfigure}
    \includegraphics[width=\textwidth,height=150pt]{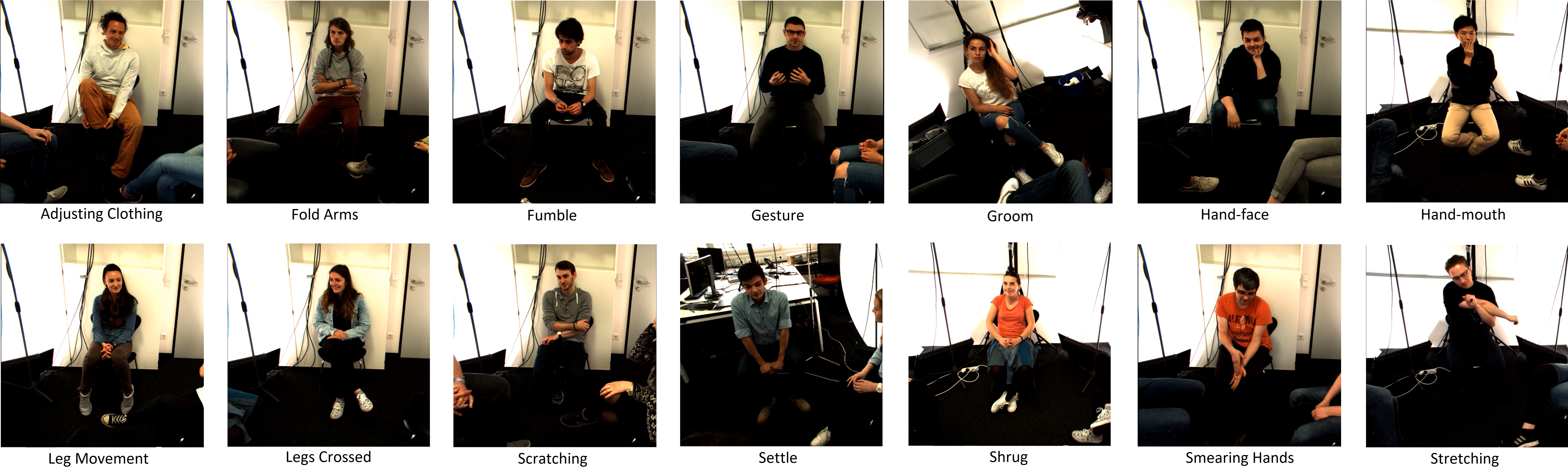}
    \caption{Examples of annotated bodily behaviors.}
    \label{fig:teaser}
\end{teaserfigure}

\maketitle

\section{Introduction}
%\philipp{Change in the whole paper: focus on time we annotated the classes, instead of number of instances per class}
%\philipp{change in intro+abstract and potentially elsewhere: analysis of connections of body language to high-level attributes on our dataset no contribution anymore}

%\philipp{change the story: annotating body behaviors is difficult. we study which behaviors can be reliably annotated, and provide a dataset of classes that can be reliably annotated.}

% part 1: body language is an important signal in social interaction
Bodily movements and poses are
a key aspect of human behavior in social interaction~\cite{vinciarelli2009social}
%that humans are even able to perceive without consciousness~\cite{tamietto2010neural}.
and are indicative of a large variety of personal and interpersonal information~\cite{izun2021}.
%and even serve as diagnostic criteria for psychiatric conditions~\cite{foley2010nonverbal}.
For example, leaning of the torso was found to be related to liking of the addressee~\cite{mehrabian1968relationship}, and behaviors like fumbling, grooming or face touching are related to the regulation of stress~\cite{bardi2011behavioral}.
Furthermore, body language was shown to have distinct effects both on its perceivers as well as on its producers.
Dominant, open bodily displays can be perceived as attractive~\cite{vacharkulksemsuk2016dominant} and gesturing was shown to lighten the cognitive load~\cite{goldin2001explaining,ping2010gesturing} and improve memory~\cite{cook2010gesturing}.
As a result, machines that are supposed to understand and participate in social interactions need to be able to accurately sense and interpret body language.

%For example, behaviors like fumbling, grooming can signal relaxation, assertion, or stress~\cite{brune2008nonverbal} and thus provide important insights into group dynamics~\cite{}.

% part 2: 
Over recent decades, huge advances were made in human body- and hand pose estimation~\cite{andriluka2008people,cao2021openpose,insafutdinov2016deepercut,ng2021body2hands}.
At the same time, a large number of works investigated the prediction of high-level attributes based on bodily behavior~\cite{beyan2017moving,muller2015emotion,visapp22}.
For example, body movements were utilized for detection of emergent leadership~\cite{beyan2017moving} and recognition of emotions~\cite{muller2015emotion} or personality types~\cite{visapp22}.
These approaches typically use generic feature sets extracted from pose estimates or rely on CNN-based visual representations.
While such approaches have the advantage of being relatively task-agnostic, they run the danger of missing subtle differences in behavior, such as between scratching and fumbling, that can only be exploited with fine-grained annotation.
They further suffer from subjective or ambiguous annotation and from the lack of interpretability associated with a psychologically-motivated mid-level representation of behavior~\cite{GIRARD2014641,OLIVEROS2015,SCHERER2014648,troisi1999ethological}, which is especially important if a behavior analysis is supposed to be accepted by practitioners like clinical or organizational psychologists.

Despite the advantages of a mid-level representation of bodily behavior in human interactions, automatic approaches for the detection of such behaviors are scarce~\cite{liu2021imigue,beyan2020analysis}.
The main reason for this is the lack of suitable datasets for training and evaluation.
The few existing datasets either only cover a single behavior like touching the face with the hands~\cite{beyan2020analysis}, or focus on single people only and are at present not publicly available~\cite{liu2021imigue}.
To overcome this limitation, we present the first publicly available annotations of a comprehensive set of body language classes embedded in continuous group conversations.
Our choice of behavior classes is motivated by previous work in psychology~\cite{troisi1999ethological}.
As a basis for annotation, we make use of a naturalistic multi-view group interaction dataset~\cite{muller2018detecting,muller2021multimediate} which will enable future research to study body language in the context of high-level social phenomena such as leadership, rapport, or liking.
%We choose the MPIIGroupInteraction dataset~\cite{muller2018detecting,muller2021multimediate} as a basis for our novel annotations, as it 
%the already existing annotations on this dataset present us with the opportunity to directly relate our new annotations with further social group phenomena.

% part 3:
Our specific contributions are threefold:
First, we introduce Bodily Behaviors in Social Interaction~(BBSI), a set of novel annotations for 15 bodily behavior classes on the MPIIGroupInteraction dataset~\cite{muller2018detecting}.
BBSI comprises 2.87 million frames of annotated behavior classes from 26 hours of human behavior embedded in continuous group interactions.
Second, we provide detailed descriptive analyses on the collected annotations as well as the results of a dedicated experiment quantifying annotator agreement.
Third, we evaluate several state-of-the-art action detection approaches on BBSI, reaching 61.3\% True Positive Rate with the Pyramid Dilated Attention Network~\cite{dai2021pdan} and Swin Transformer~\cite{liu2021Swin} features.\footnote{Data and code are available at \url{https://git.opendfki.de/body_language/acm_mm22}.}

\section{Related Work}
Our work is related to the function of body language in social interactions, to approaches for the recognition of actions and body language, as well as to existing human body language datasets.

\subsection{Body Language in Social Interaction}
\label{sec:classes}

Body language has been actively researched by psychologists for decades~\cite{mehrabian1968relationship,goldin2001explaining,vacharkulksemsuk2016dominant}. Early work by Mehrabian~\cite{mehrabian1968relationship,mehrabian1969encoding} found that, among other signals, backward leaning of the torso is indicative of liking.
%Not only does a persons' body language tell about their attitude towards the interlocutor, but it also influences the interlocutors' attitude toward that person.
Dominant and open nonverbal displays, as opposed to folded arms and crossed legs, are perceived as attractive when meeting with strangers~\cite{vacharkulksemsuk2016dominant}. In a meta-analysis, \cite{hall2005} found significant correlations between perceived social verticality and for example self-touching and gesturing.  A further study by~\cite{carney2005} indicated that people believe power is expressed with nonverbal cues like open posture (i.e. no arms crossed or legs crossed), more gesturing, and less self-touching (both hands and face). Furthermore, leaning towards the interlocutor was shown to be associated with rapport~\cite{sharpley1995}, and crossed arms were shown to be associated with emotion expressions~\cite{wallbott1998}. Displacement behaviors such as grooming, face touching or fumbling  are related to anxiety and stress regulation~\cite{bardi2011behavioral,mohiyeddini2015neuroticism,mohiyeddini2013displacement}. 
%Gesturing when talking was shown to lighten cognitive load~\cite{goldin2001explaining,ping2010gesturing}, to improve memory~\cite{cook2010gesturing}, and to be predictive of domain expertise~\cite{sriramulu2019dynamic}.

As a consequence of these manifold connections of body language with important personal and social attributes, body language analysis has been a focus of automatic approaches attempting to infer high-level attributes such as emotion~\cite{muller2015emotion,gavrilescu2015recognizing,ranganathan2016multimodal}, leadership role~\cite{beyan2017moving,muller2019emergent}, or personality type~\cite{visapp22,romeo2021predicting}. In contrast to 
%the approach employed by 
the human science studies discussed above, these automatic approaches commonly lack an explicit intermediate representation of functional bodily behavior categories. 
Instead, they rely on a generic feature representation encoding body postures and movements~\cite{muller2015emotion,beyan2017moving,muller2019emergent} or on deep learning approaches~\cite{romeo2021predicting,ranganathan2016multimodal} without easily interpretable internal structure. 
While such representations can be effective in prediction scenarios, they often lack interpretability and may miss subtle but meaningful differences, e.g. between fumbling and scratching. 
%While such representations can be effective in prediction scenarios, they are often hard to explain and may also suffer from misunderstandings of subtle but meaningful differences, e.g. between fumbling and scratching. 
In this work, we draw upon the ethological rating scheme of functional body language categories described in~\cite{troisi1999ethological} to derive a set of bodily behaviors that are intuitively interpretable and allow to train models for fine-grained behavior distinctions.

%\philipp{In the camera ready version, we will explicitly include the phenomena leadership, rapport, and emotion. For example, concerning the connections of our behavior classes to social verticality - a concept closely related to emergent leadership, we will discuss the meta-analysis of~\cite{hall2005}. This meta-analysis reported significant correlations between perceived social verticality and for example self-touching and gesturing, which are also represented in our behavior classes. A further study by~\cite{carney2005} indicated that people believe power is expressed with nonverbal cues like open posture (i.e. no arms crossed or legs crossed), more gesturing, and less self-touching (both hands and face). Furthermore, leaning towards the interlocutor was shown to be associated with rapport~\cite{sharpley1995}, and crossed arms were shown to be associated with emotion expressions~\cite{wallbott1998}.}

\begin{table*}[t]
    %\vspace{-5pt} %%%
    \caption{Datasets annotated with body language classes described in the literature. \textit{Behaviors} indicates the number of annotated body language classes, \textit{Participants} the number of human individuals, \textit{Length} the length of annotated behavior, \textit{Views} the number of synchronized camera views on each participant, \textit{Group Size} the number of participants that were synchronously annotated, \textit{Spontaneous} whether behavior was shown spontaneously, and \textit{Public} whether the dataset is publicly available.
    NTU is in italic, as only a subset of its classes are body language.}
    \vspace{-10pt} %%%
    \begin{tabular}{l | rrrrrrr } 
    \toprule
    Name & Behaviors & Participants & Length & Views & Group Size & Spontaneous & Public \\
    \midrule
    iMiGUE~\cite{liu2021imigue} & 32 & 72 & 35h & 1 & 1 & \cmark & \xmark\\
    PAVIS Face-Touching~\cite{beyan2020analysis} & 1 & 64 & 22h & 1 & 4 & \cmark & \cmark \\
%    HUMAINE~\cite{douglas2007humaine} & 8 & 10 & 1 & 1 & \xmark &  \cmark \\
    EMILYA~\cite{fourati2014emilya} & 7 & 11 & 6h & 1 & 1 & \xmark & \cmark \\ 
%    LIRIS-ACCEDE~\cite{gavrilescu2015recognizing} & 6 & 64 & 1 & 1 & \xmark &  \cmark \\ 
%    emoFBVP~\cite{ranganathan2016multimodal} & 23 & 10 & 1 & 1 & \xmark &  \cmark \\
    \textit{NTU RGB+D 60/120~\cite{ntu2016,ntu2020}} & \textit{60/120} & \textit{40/106} & \textit{133h/266h} & \textit{80/155} & \textit{1--2} & \xmark & \cmark \\ 
    \midrule
    BBSI (ours) & 15 & 78 & 26h & 3 & 3--4 & \cmark & \cmark \\
    \bottomrule
    \end{tabular}
    %\vspace{-5pt} %%%
    \label{tab:datasets}
\end{table*}

\subsection{Recognition of Actions and Body Language}

RGB-based human action recognition has often been addressed by three main approaches. Two-stream 2D Convolutional Neural Networks~\cite{ziss2014,KarpathyCVPR14,ZONG2021104108} generally contain two 2D CNN branches taking different input features extracted from the RGB videos for action recognition. Recurrent Neural Networks~(RNN)~\cite{liu2020,Ng_2015_CVPR,7558228} usually employ 2D CNNs as feature extractors for an LSTM model. 3D CNN-based methods~\cite{Tran2018ACL,9093274,9578601} extend 2D CNNs to 3D structures, to simultaneously model the spatial and temporal context information in videos that is crucial for action recognition.

Among the many available human action recognition methods we choose the following three for our evaluations: A well-cited two-stream 2D CNN architecture by Wang et al.~\cite{TSN2016ECCV} which divides each video into three segments and processes each segment with a two-stream network, fusing the individual classification scores by an average pooling method to produce the video-level prediction. A revolutionary method by Carreira and Zisserman~\cite{8099985} which introduces the two-stream Inflated 3D CNN inflating the convolutional and pooling kernels of a 2D CNN with an additional temporal dimension. And the best performance method tested by~\cite{liu2021imigue} on body language recognition by Lin et al.~\cite{lin2019tsm} of a parameter-free Temporal Shift Module, which shifts a part of the channels along the temporal dimension to perform temporal interaction between the features from adjacent frames. We also experiment with the transformer method by Liu et al.~\cite{liu2021Swin} that was designed for natural language processing but its application has been recently extended to computer vision tasks~\cite{dosovitskiy2021an,khan2021}.

%In contrast to action recognition, 
%the work on the related problem body language recognition is much thinner.
%%, even though the two problems are closely related. 
%However, most of the existing works on action recognition are designed for scenarios in which people move rather freely and are engaged in variable activities. 
%These methods are targeting for the actions annotated in general action datasets such as Kinetics~\cite{kinetics}, Smarthome~\cite{Das_2019_ICCV} or Charades~\cite{Sigurdsson2016HollywoodIH}. Limiting to the social interaction environment brings a large set of constraints including restricted motion across the scene, compact body poses or subtle facial expressions and gestures.

In contrast to action recognition, which typically considers freely moving people ~\cite{kinetics,Das_2019_ICCV,Sigurdsson2016HollywoodIH}, the much thinner work on body language recognition addresses more constrained social interaction scenarios.
%, even though the two problems are closely related. 
%However, most of the existing works on action recognition are designed for scenarios in which people move rather freely and are engaged in variable activities. 
%These methods are targeting for the actions annotated in general action datasets such as Kinetics~\cite{kinetics}, Smarthome~\cite{Das_2019_ICCV} or Charades~\cite{Sigurdsson2016HollywoodIH}. Limiting to the social interaction environment brings a large set of constraints including restricted motion across the scene, compact body poses or subtle facial expressions and gestures.
For example, Yang et al.~\cite{yang2020} generate sequences of body language predictions from estimated human poses and feed them to an RNN for emotion interpretation and psychiatric symptom prediction. Kratimenos et al.~\cite{Kratimenos2021IndependentSL} extract a holistic 3D body shape, including hands and face, from a single image and feed them also to an RNN for sign language recognition. Singh et al.~\cite{7375234} use hand-crafted features to analyze body language for estimating a person's emotions and state of mind. Santhoshkumar et al.~\cite{SANTHOSHKUMAR2019158} use Feedforward Deep CNNs for detecting emotions from full body motions. 
We observe that the common denominator of body language analysis methods are the employment of a general action recognition method and the lack of a benchmark body language dataset.

\subsection{Human Body Language Datasets}

In contrast to datasets with annotations of high-level attributes like emotions~\cite{ranganathan2016multimodal,muller2015emotion}, leadership~\cite{beyan2017moving,muller2018detecting}, or personality~\cite{palmero2021context}, datasets annotated with concrete classes of bodily behavior are sparse.
Table~\ref{tab:datasets} summarizes four relevant datasets with manual body language annotations.
Research that extracted body language automatically but did not provide human annotations is not included~\cite{gavrilescu2015recognizing,ranganathan2016multimodal,douglas2007humaine}.
%Most of these datasets use posed expressions~\cite{douglas2007humaine,fourati2014emilya,gavrilescu2015recognizing}.

In EMILYA~\cite{fourati2014emilya}, actors were asked to express different emotions while performing daily actions such as walking, sitting down, or moving objects. 
%While acted expressions are a convenient means to create a dataset, the ecological validity of such an approach 
%HUMAINE consists of \philipp{explain}
Two datasets NTU RGB+D 60/120~\cite{ntu2016,ntu2020} contain a large number of general action classes that also include a number of body language classes.
%They even include the depth dimension and capture the scene from impressive 80/155 viewpoints. 
However, these datasets do not consist of spontaneous human behavior.
The two most relevant datasets for our work are the PAVIS Face-Touching dataset~\cite{beyan2020analysis} and iMiGUE~\cite{liu2021imigue}.
PAVIS Face-Touching is similar to BBSI as it also consists of recordings of group discussions.
In contrast to the 15 behavior classes annotated in BBSI, PAVIS Face-Touching only has binary annotations of whether a participant touches her face or not.
Furthermore it only has a single frontal view on each participant.
The recently introduced iMiGUE dataset~\cite{liu2021imigue} consists of annotations of 32 behaviors classes of speakers at sports press conferences.
Annotations are only provided for a single person (i.e. no annotations of discussion partners), and only a single view on the target person is provided.
At the time of submission, the iMiGUE videos are not publicly accessible due to privacy issues\footnote{According to a note dating from September 2021 on the official github page of iMiGUE (\url{https://github.com/linuxsino/iMiGUE}), the file containing the links to the videos used in the dataset has been removed for privacy protection.}.
We hereby present the first publicly available annotations of body language on a multi-view dataset of three to four people engaged in spontaneous group discussions.

\section{Dataset}
% 1: Adjusting Clothing, 2: Fold arms, 3: Fumble, 4: Gesture, 5: Groom, 6: Hand-face, 7: Hand-mouth, 8: Lean towards, 9: Leg Movement, 10: Legs crossed, 11: Scratch, 12: Settle, 13: Shrug, 14: Smearing Hands, 15: Stretching

BBSI builds upon the MPIIGroupInteraction dataset~\cite{muller2018detecting}.
This dataset comprises of 22 three- to four-person group discussion on controversial topics, each lasting for 20 minutes.
In total, it consists of 78 participants and 26 hours of behavior recordings.
Every interaction was recorded by 8 frame-synchronized cameras as well as with 4 microphones.
After the discussions, participants rated their perceived leadership, competence, dominance and liking of all other members, as well as their feelings of rapport towards each other.
In addition to rapport and emergent leadership prediction~\cite{muller2018detecting,muller2019emergent}, the dataset was further annotated and used for eye contact detection~\cite{muller2018robust,fu2021using} and for next speaker prediction~\cite{muller2021multimediate,birmingham2021group}.
This wealth of already existing annotations makes the MPIIGroupInteraction dataset a perfect choice for the collection of body language labels as it will allow future research on the connections and the utility of body language information with key group phenomena.
%Adding body language information to the wealth of already existing annotations on the MPIIGroupInteraction dataset creates opportunities.

\begin{table*}[t]
    \vspace{-5pt} %%%
    \caption{Behavior classes in the dataset, including descriptions, number of annotated frames, annotation instances, and annotator agreement.}
    \vspace{-10pt} %%%
    \begin{tabular}{l|p{8.3cm}rrr} 
    \toprule
    Behavior & Description & \# Frames & \# Instances  & Agreement \\
    \midrule
    Adjusting Clothing & Clothing is adjusted & 23k & 250 & 0.77 \\
    Fold Arms & Arms are folded across the chest & 251k & 200 & 0.82\\
    Fumble & Twisting and fiddling finger movements & 422k & 1374 & 0.54\\
    Gesture & Variable hand and arm movements during speech & 373k & 2607 & 0.85\\
    Groom & Fingers are passed through the hair in a combing movement & 17k & 282 & 0.71\\
    Hand-face & Hand(s) in contact with the face & 79k & 535 & 0.79\\
    Hand-mouth & Hand(s) in contact with the mouth  & 55k & 318 & 0.74\\
    Lean Towards & Leaning forward from the hips towards the interlocutor & 5k & 72 & 0.13\\
    Leg Movement & Repetitive movement of legs & 14k & 860 & 0.51\\
    Legs Crossed & Legs are crossed & 1397k & 77 & 0.87\\
    Scratch & Fingernails are used to scratch parts of the body & 72k  & 519& 0.61\\
    Settle & Adjusting movement into a more comfortable posture in the chair & 40k & 290 & 0.54\\
    Shrug & Shoulders are raised and dropped again & 8k  & 192 & 0.57\\
    Smearing Hands & Smearing hands on clothing & 21k & 298 & 0.54\\
    Stretching & Stretching of body parts & 4k & 31 & 0.61\\
    \bottomrule
    \end{tabular}
    \label{tab:behavior_classes}
    \vspace{-6pt} %%%
\end{table*}

\subsection{Body Language Annotation}

We densely annotated the full MPIIGroupInteraction dataset with 15 body language classes~(see Figure~\ref{fig:teaser} and Table~\ref{tab:behavior_classes}).
Our set of behavior classes is based on the the Ethological Coding System for Interviews~(ECSI)~\cite{troisi1999ethological}.
This coding system includes many bodily behaviors that were shown to be connected to different social phenomena, as described in Section~\ref{sec:classes}.
We selected all ECSI behaviors involving the limbs and torso and excluded behavior classes based on facial behavior, gaze, and head pose as these are not the focus of this work and highly accurate methods to analyze such behaviors already exist~\cite{baltrusaitis2018openface,sinha2021flame}.
We also excluded the two classes \textit{Crouch} and \textit{Relax}, as they were only very rarely annotated~(Crouch: 411 frames, Relax: 2k frames), rendering estimation of classification performance meaningless.
In addition to the bodily behaviors included in ECSI, we scanned the MPIIGroupInteraction dataset for additional behaviors that occur frequently and carry potential meaning in a social situation.
As a result, we included the five additional classes: \textit{Adjusting Clothing}, \textit{Leg Movement}, \textit{Legs Crossed}, \textit{Smearing Hands}, \textit{Stretching}.

To achieve high-quality annotations while keeping costs manageable, we designed the following annotation procedure.
First, we trained three annotators on the task by providing examples and discussing edge cases jointly.
In this way, we made sure that the annotators arrived at a common understanding of the body language classes.
Each of the 78 participants of the MPIIGroupInteraction dataset was fully annotated by one of the annotators.
Subsequently, each of the resulting annotations was checked by another annotator to further improve quality.
This procedure of annotation followed by checking proved to be much more economical than collecting several separate annotations of the same video.
We used a separate experiment to quantify annotation quality~(see Section~\ref{sec:dataset_analysis}).

\subsection{Analysis of Annotations}
\label{sec:dataset_analysis}

\subsubsection{Descriptive Statistics}

In total, 2.87 million frames of body language were annotated across all classes for the full 26 hours of video.
Each annotation instance is defined by a specific behavior label and a start time and an end time between which the behavior appears continuously on all frames. Table~\ref{tab:behavior_classes} shows that the annotation across the 15 behavior classes has highly uneven number of annotated frames and instances.
The most frequently annotated class, \textit{Legs Crossed} was annotated for 1397k frames, while \textit{Stretching} was only annotated for 4k frames.
As a complementary view on the quantity of annotations, \textit{Legs Crossed} has the highest number of annotated frames, but only over 77 annotation instances, meaning that participants remained for a crossed-leg position for extended periods of time.
On the other hand, \textit{Gesture} is annotated on less frames (373k), but consists of many more distinct instances (2607).

\begin{figure*}[t]
    \vspace{-5pt} %%%
    \includegraphics[width=0.65\textwidth]{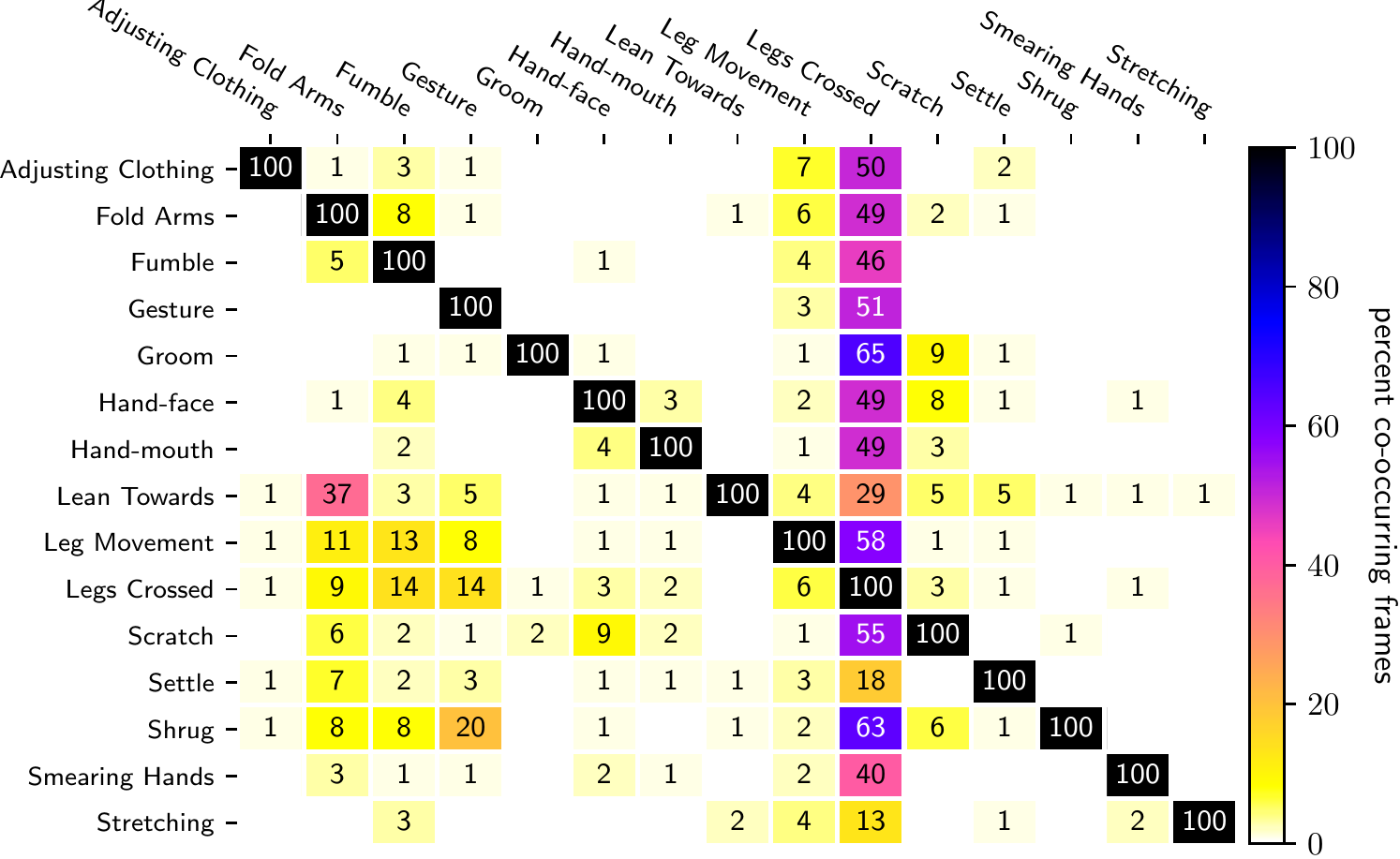}
    \includegraphics[width=0.34\textwidth]{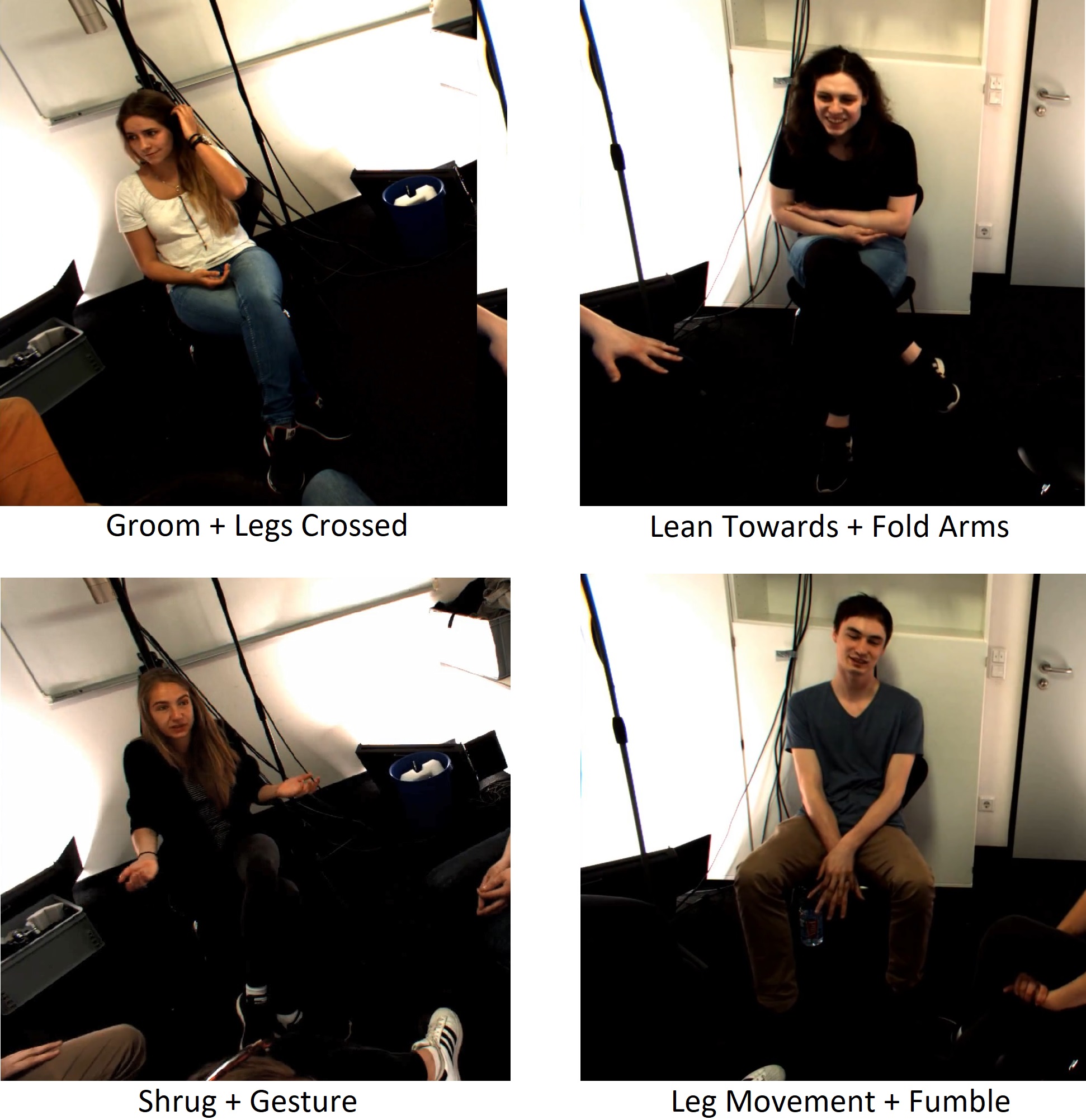}
    \vspace{-10pt} %%%
    \caption{Co-occurrences of body language classes. Each row shows the percentage with which other classes are annotated at the same time as the class designated on the y-axis.}
    \label{fig:class_cooccurrences}
    %\vspace{-5pt} %%%
\end{figure*}

Another important aspect of BBSI is its multi-label characteristic, that is, several body language classes can occur simultaneously.
Figure~\ref{fig:class_cooccurrences} shows the co-occurrence patterns of body language classes.
Strong co-occurrences can be observed between the lower body classes~(\textit{Legs Crossed}, \textit{Leg Movement}) and upper-body behaviors.
Co-occurrences between upper-body behaviors do exist, but are more sparse.
As a result, BBSI creates a challenging multi-label classification problem.

\subsubsection{Annotation Quality}

To obtain a numerical estimate of annotation quality, we performed a dedicated experiment based on the collected annotations.
We sampled 800 4-second clips from the full dataset that were classified into body language classes separately by all three annotators.
These samples were drawn randomly from the whole dataset with the following constraints:
%These samples were drawn randomly from the whole dataset subject to the following constraints:
First, we considered a 4-second window to be a sample of a body language class if either the class is annotated for at least 2 seconds of this window, or if the 4-second window completely encompasses the corresponding annotation instance.
%First, we considered a 4-second window to be a sample of a body language class if either the class is annotated for at least 2 seconds of the 4 second window, or if the 4-second window completely encompasses the corresponding annotation instance.
Second, we drew 50 samples of each behavior class.
Estimated with the rate of class co-occurrences, the precise number of instances for each class in the 800 samples may be larger than 50.
For comparability, we used the same metric as~\cite{liu2021imigue} which computes the agreement of two annotators by dividing twice the number of annotated behaviors for which they agree by the total number of behaviors annotated by both.

Table~\ref{tab:behavior_classes} shows the resulting agreements for each class separately.
Very high agreements above 0.8 are reached for frequent classes such as \textit{Legs Crossed}, \textit{Gesture}, or \textit{Fold Arms}.
All other classes are in the range of 0.5 to 0.8 with the only exception of \textit{Lean Towards} which was proven very challenging to annotate with only 0.13 agreement.
Liu et al.~\cite{liu2021imigue} do not provide class-specific annotator agreement but only a global measure in which frequent classes contribute more than less frequent classes, i.e. micro average. 
%Liu et al.~\cite{liu2021imigue} do not provide class-specific annotator agreement but only a global measure in which frequent classes contribute more than less frequent classes, that is, the micro average. 
To get an estimate how our annotator agreement relates to the agreement of 0.81 reported in~\cite{liu2021imigue}, we weight our class-specific agreements by the frame-wise label distribution on BBSI, reaching an agreement of 0.78.
Note however, that these numbers are not directly comparable due to different behavior classes and annotation protocols.

%\subsubsection{Connection to Speaking Turns}
%To investigate 

%\subsubsection{Connection to High-Level Attributes}
%We investigate the connections to 
%Figure~\ref{fig:bl_to_high_level}
%
%\begin{figure}[h]
%    \includegraphics[width=0.49\columnwidth]{fig/Duration_PLead.pdf}
%    \includegraphics[width=0.49\columnwidth]{fig/Duration_rapport.pdf}
%    \caption{Correlations between the the frequency of body language classes shown by participants and received leadership (left) and rapport ratings (right). The dotted line indicates significance of the correlation coefficient at $\alpha = 0.05$ (not corrected for multiple tests). \philipp{add -0.5 on y axis} \michal{or -0.3 if it looks better. You can also add a 3rd plot and shrink everything horizontally.}}
%    \label{fig:bl_to_high_level}
%\end{figure}

%\begin{itemize}
%    \item one good example for each behavior class 
%    \item examples showing high intra-class variability / challenges
%\end{itemize}

%\begin{figure}[h]
%    \includegraphics[width=0.49\columnwidth]{fig/anno_counts}
%    \includegraphics[width=0.49\columnwidth]{fig/anno_durs}
%    \caption{Annotations collected per each behavior class in terms of count~(left) and in terms of time~(right). \michal{Can be merged in one column plot with two columns per class and second axis on the right side.} \philipp{I'd propose to integrate these numbers into Table 1. This will save quite some space.} \michal{I agree!}}
%    \label{fig:anno_counts}
%\end{figure}

\section{Method}
For detecting the behaviors in the long input videos, we propose a baseline method based on the Pyramid Dilated Attention Network~\cite{dai2021pdan} for action detection. The model is fed with features extracted by four types of action recognition architectures. 

\subsection{Feature Extraction Networks}

%Features are extracted by first training an action recognition model as a classifier on the training part of our dataset, followed by removing its last classification layer to serve as a feature extractor and inferring feature extraction from the whole dataset. 
We examine the following four established algorithms that are designed for general action recognition tasks.

\subsubsection{Two-Stream Inflated 3D CNN}
Extent of a pre-training boost depends on the ability of a model architecture to adapt to a given pre-training dataset. As the 3D image classification backbone, the Two-Stream Inflated 3D CNN~(I3D)~\cite{8099985} uses the ImageNet-pretrained Inception V1 with batch normalization. Filters and pooling kernels of very deep 2D image classification CNNs are inflated into 3D to learn spatio-temporal feature extractors from video.

\subsubsection{Temporal Segment Networks}
An obvious problem of the two-stream CNNs is their inability to model long-range temporal structure due to their access to only a limited stack of frames. Temporal Segment Networks~(TSN)~\cite{TSN2016ECCV} operate on a sequence of short video clips sparsely sampled from the entire video. Each clip in this sequence will produce its own preliminary prediction of the action classes. Prediction over the full video is then derived from a consensus among the partial clip predictions. In the learning process, the loss values of video-level predictions, other than those of clip-level predictions which were used in two-stream CNNs, are optimized by iteratively updating the model parameters.

\subsubsection{Temporal Shift Module}
Traditional 3D convolution uses a 3D convolution kernel to perform convolution operations between adjacent multiple frames at the same time, which can extract the spatio-temporal feature information in the video at the cost of an increase in calculation. Temporal Shift Module~(TSM)~\cite{lin2019tsm} uses a simple data preprocessing method to convert the invisible temporal information in a single frame into extractable spatial feature information. Several consecutive frames are stacked to form the original tensor and the channels are moved forward and backward in the temporal dimension to perform a simple feature fusion between the consecutive frames. The fusion makes an independent single frame contain certain temporal information, and simple 2D convolution can be used to achieve spatiotemporal feature extraction.

\subsubsection{Swin Transformers}
Adapting the network architectures in natural language processing to the domain of computer vision suffers from large variations in the scale of visual entities and the high resolution of pixels in images compared to words in text. Building upon the Transformer designed for sequence modeling and translation tasks, the Swin Transformer~(Swin)~\cite{liu2021Swin} is a hierarchical Transformer whose representation is computed with Shifted windows. The shifted windowing scheme brings greater efficiency by limiting self-attention computation to non-overlapping local windows while also allowing for cross-window connection.

\subsection{Training Feature Extractors}

As human body language ground truth contains temporally overlapping labeled segments as well as unlabeled sections, we investigate two training settings for the feature extraction networks: \textit{Single-Label} training and \textit{Multi-Label} training. In Multi-Label training, a label is assigned to a clip if it overlaps an annotated segment in at least half of its duration. 
On BBSI, 25\% of the samples have no label assigned, 49\% samples have one label and the remaining 26\% on an overlap have between two and five labels. 
For the Single-Label setting, we selected the samples from the Multi-Label setting with at least one label.
Each sample was assigned precisely one label and those originally with multiple labels are copied multiple times, each time with a single and unique label.
The loss function we used was the cross entropy loss followed by the softmax activation function.
For each sample, each feature extractor returns a confidence score for each behavior class. 
Prior to the output layer, each feature extractor produces a 2048-dimensional feature vector that is used as input to the behavior detection method. 

\subsection{Behavior Detection}

Detection of behaviors from a long video is done by feeding the extracted feature vectors into an action detection architecture.
The Pyramid Dilated Attention Network~(PDAN)~\cite{dai2021pdan} uses a self-attention mechanism to capture temporal relations. The layers that make up the network are called Dilated Attention Layers~(DAL). A DAL takes each segment as a center segment and concatenates its feature representation with the feature representation of segments being $D$-far in both directions from the center segment, where $D$ is the dilation rate. At this point, it applies self-attention on the extracted segment-feature representations. PDAN is based on a pyramid of DALs with same kernel sizes and dilation rates that exponentially increase their temporal receptive field.
The output of PDAN consists of a list of predicted behaviors with their beginnings and endings tied to the segmentation cuts, and their confidences. 

%\begin{figure}[h]
%    \includegraphics[width=0.8\textwidth]{fig/pdan.png}
%    \caption{Overview of the Pyramid Dilated Attention Network~(PDAN)~\cite{dai2021pdan}. \michal{Redraw.}}
%    \label{fig:pdan}
%\end{figure}

\subsection{Implementation Details}

The feature extraction methods operate on fixed inputs of length 16 frames and size $224\!\times\!224$ pixels. 
Consequently, we resize the videos appropriately and cut the long dataset videos into 16-frame video clips. 
%To make use of these methods, we resize the videos accordingly and cut the long dataset videos into short 16-frame video clips. 
These clips are assigned with the corresponding behavior class label and treated as independent samples for training and evaluation.
%Assigned with the annotated behavior class labels, these clips are treated as independent samples for training and evaluation. 
This splitting of videos does not disconnect the flow of the actions as the annotated behaviors are mostly non-transitional, that is, the actions described by these behaviors do not change people's body poses from one to another. 
%Splitting the videos this way does not disconnect the flow of the actions as the annotated behaviors are mostly non-transitional, that is, the actions described by these behaviors do not change people's body poses from one to another.
For instance, a 64-frame-long behavior \textit{Hand-mouth} can be split into four 16-frame-long clips in which a person keeps touching their mouth. 
Advantages are that the number of samples increases significantly and that the fixed-length clips can be input to all methods with an equal FPS.
%Two notable advantages are that the number of samples increases significantly and that the fixed-length clips can be input to all methods with an equal FPS.

All action recognition models are pre-trained on ImageNet and Kinetics-400 and the action detection model is used without any pre-training. Fine-tuning on BBSI is performed on both levels, action recognition and action detection.
%It contains the implementations and pre-trained models of many well-known action recognition architectures and supports both labeling settings.
For comparability, all models were trained for 15 epochs.
Learning rates are set to: I3D $10^{-2}$, TSN $10^{-3}$, TSM $7.5\cdot10^{-4}$, Swin $10^{-3}$ with AdamW optimizer, and PDAN $10^{-1}$.
Our implementation uses the open-source toolbox MMaction2~\cite{2020mmaction2} built on top of PyCharm. 
%Trained models are subsequently stripped of their final classification layers and used as feature extractors. 
%In addition to evaluation of behavior detection, the following section provides a quality analysis of the extracted features.

\section{Evaluation}
We provide evaluation results of our baseline method at two levels: quality of the extracted features and quality of the final detection. 
%In this section we provide evaluation results of our baseline method at two levels: quality of the extracted features and quality of the final detection.
As feature extractors are trained as classifiers, they are evaluated with standard classification metrics, and the final detector is evaluated on standard detection metrics.
%As feature extractors are trained as classifiers, they are evaluated in terms of standard classification metrics, and the final detector is evaluated on standard detection metrics.
In all experiments, we use the training/validation split of MPIIGroupInteraction reported in \cite{muller2021multimediate}: training recordings 07, 10--25; validation recordings 08, 09, 26--28.

\begin{table}[t]
    %\vspace{-5pt} %%%
    \caption{Five action recognition architectures, four proposed methods and one random, in both labeling settings. Each value is MAP score computed using micro/macro averaging.}
    \vspace{-10pt} %%%
    \begin{tabular}{l|cc}
    \toprule
    Method & Single-Label & Multi-Label \\
    \midrule
    random & 0.258 / 0.067 & 0.377 / 0.106 \\
    I3D~\cite{8099985} & 0.445 / 0.212 & 0.624 / 0.284 \\
    TSN~\cite{TSN2016ECCV} & 0.520 / 0.232 & 0.661 / 0.308 \\
    TSM~\cite{lin2019tsm} & 0.508 / 0.228 & 0.721 / 0.313 \\
    Swin~\cite{liu2021Swin} & 0.601 / 0.305 & 0.745 / 0.374 \\
    \bottomrule
    \end{tabular}
    \label{tab:map_all}
    \vspace{-5pt} %%%
\end{table}

\begin{figure}[t]
    \vspace{-5pt} %%%
    \subfigure[Single-Label Swin~\cite{liu2021Swin}]{\includegraphics[width=0.475\textwidth]{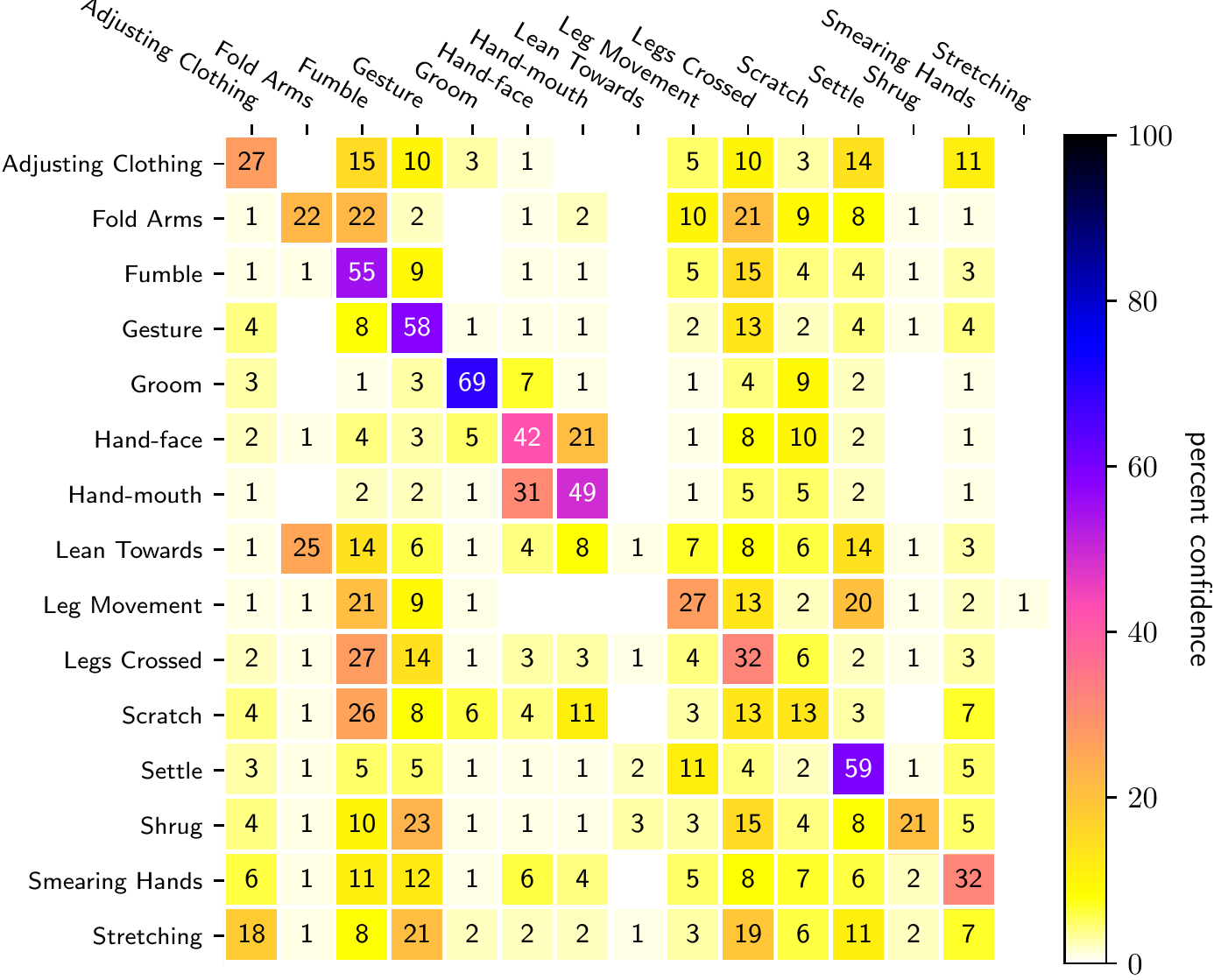}}%
    \vspace{-5pt} %%%
    \subfigure[Multi-Label Swin~\cite{liu2021Swin}]{\includegraphics[width=0.475\textwidth]{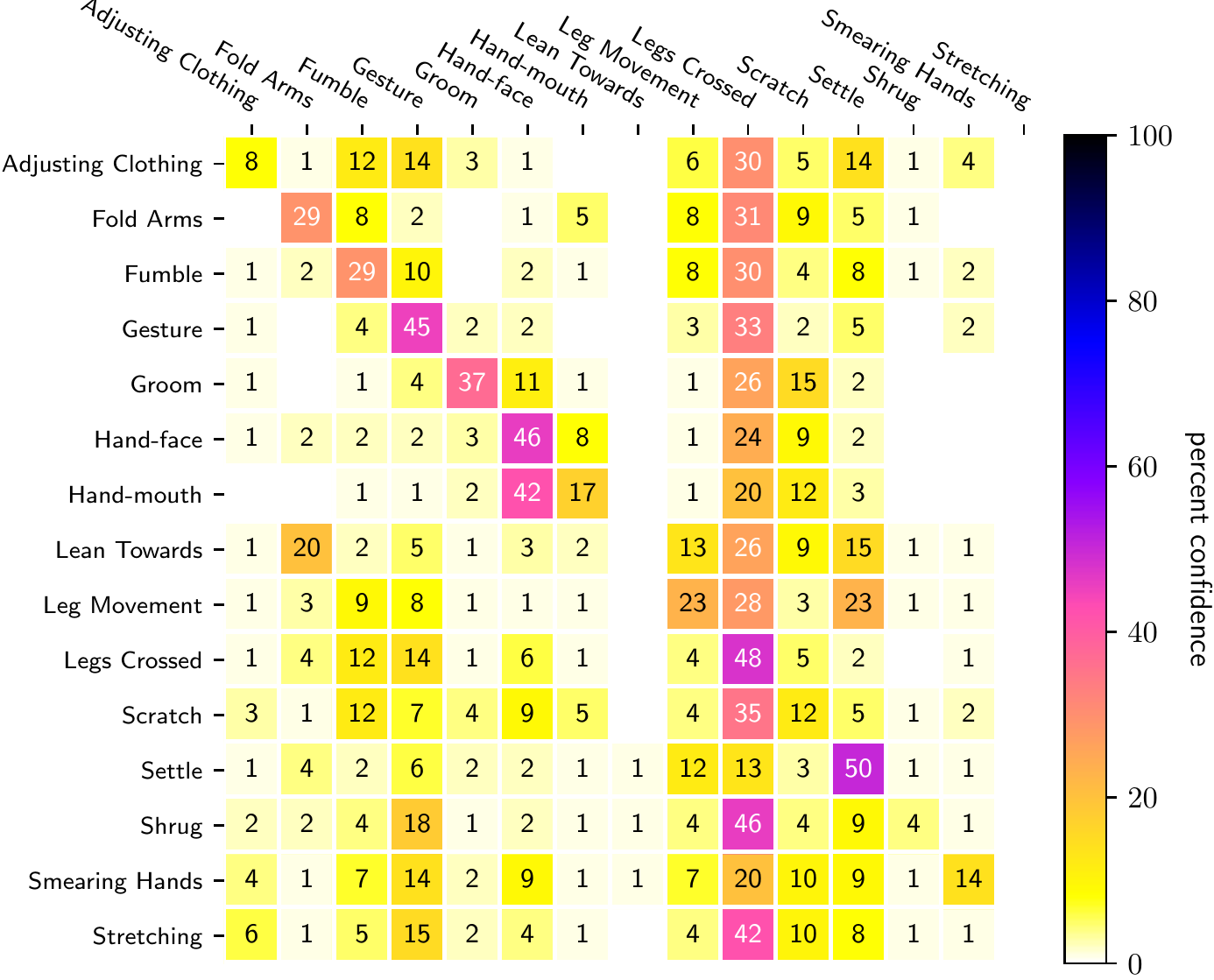}}%
    \vspace{-5pt} %%%
    \caption{Confidence matrices of behavior recognition by the Swin Transformer trained in both labeling settings.}
    \label{fig:conf}
    \vspace{-5pt} %%%
\end{figure}

\subsection{Classification Evaluation}

Table~\ref{tab:map_all} shows classification performance of the feature extraction networks, both for the Multi-Label as well as for the Single-Label training setting. We also include the random classifier as a baseline. 
Results are reported in terms of Mean Average Precision~(MAP) using micro averaging (same weight for each sample) and macro averaging (same weight for each class).
The best result is achieved by Swin Transformers in the multi-label training scenario, reaching 0.75 micro averaging MAP and 0.37 macro averaging MAP.
The second best method in the Multi-Label setting was TSM with 0.72 and 0.31 micro- and macro averaging MAP.
In the Single-Label scenario, Swin Transformers also reached the best performance.
The second best method in this case is TSN.
%the best performance in all evaluation settings.
%both the Single-Label (0.60/0.31 MAP) as well as the Multi-Label setting.
%the is apparently dominating, the differences between the RGB-based methods are less clear.
%Interestingly, TSN has a slight advantage in the Single-Label setting and the worst results on Multi-Label at the same time.
%In the Multi-Label setting, TSM reaches 0.72 micro average and I3D 0.33 macro average, but they still fall behind Swin with 0.75 and 0.37, respectively.
All feature extraction networks clearly outperformed the random baseline.

Evaluation of all methods can be visualized by aggregating all confidence vectors into a confidence matrix. Rows of this $15\!\times\!15$ square matrix are ground truth classes and columns are prediction confidences. The matrix is constructed by adding all confidence vectors into the corresponding ground truth rows and then dividing each row by the number of its summands. 
In the ideal case, this matrix would coincide with the co-occurrence matrix presented in Figure~\ref{fig:class_cooccurrences}.
%Diagonal values are desired high as it is intended to have a high confidence in correct classes, and the remaining values need to be low as wrong classes should have a low confidence. 
%\akos{In Single-Label classification tasks the visual interpretation of the confidence matrix is straightforward: the diagonal values are desired high as it is intended to have a high confidence in correct classes, and the remaining values need to be low as wrong classes should have a low confidence. } 
%The matrix can be multiplied by 100 to represent the confusion percentages. 
%\akos{should we call it confusion?}
See Figure~\ref{fig:conf} for the confidence matrices of behavior recognition by the Swin Transformer for both Single-Label and Multi-Label training.
Compared to Single-Label training, the Multi-Label network is able to more accurately model class co-occurrences, especially with \textit{Legs Crossed}.
We report further confidence matrices in the supplementary material.

We performed additional ablation experiments with the TSM model. First, as the behaviors \textit{Legs Crossed} and \textit{Fold Arms} make forms of static positional body pose rather than dynamic motion actions, they can be recognized on a frame level and with an eventual aid of a generally non-temporal skeleton estimation technique. We evaluated the training and evaluation scenario with only 13 classes, excluding these two static classes. And second, as the dataset has considerably imbalanced class frequencies (ranging from 4k annotated frames to more than a million, see Table~\ref{tab:behavior_classes}), overrepresented behaviors have a too high impact on training compared to underrepresented ones. 
Therefore, we evaluated the influence of class balancing by randomly selecting 20k samples from each class to counteract weight of overrepresented classes while keeping all samples of the underrepresented classes. 
Table~\ref{tab:map_tsm} shows the effects of static class exclusion and class balancing. We observe systematic advantage of excluding static classes in the Single-Label setting and of including static classes in the Multi-Label setting, and of no class balancing overall.

\begin{table}[t]
    %\vspace{-5pt} %%%
    \caption{Effects of exclusion of static classes and class balancing on behavior recognition by TSM~\cite{lin2019tsm}. Each value is a MAP using micro/macro averaging.}
    \vspace{-10pt} %%%
    \begin{tabular}{cc|cc}
    \toprule
    static included & balancing & Single-Label & Multi-Label \\
    \midrule
    \cmark & \cmark & 0.508 / 0.228 & 0.721 / 0.319 \\
    \cmark & \xmark & 0.595 / 0.294 & 0.746 / 0.384 \\
    \xmark & \cmark & 0.601 / 0.279 & 0.618 / 0.305 \\
    \xmark & \xmark & 0.658 / 0.333 & 0.639 / 0.336 \\
    \bottomrule
    \end{tabular}
    \label{tab:map_tsm}
    %\vspace{-5pt} %%%
\end{table}

\begin{figure}[t]
    %\vspace{-5pt} %%%
    \includegraphics[width=0.35\textwidth]{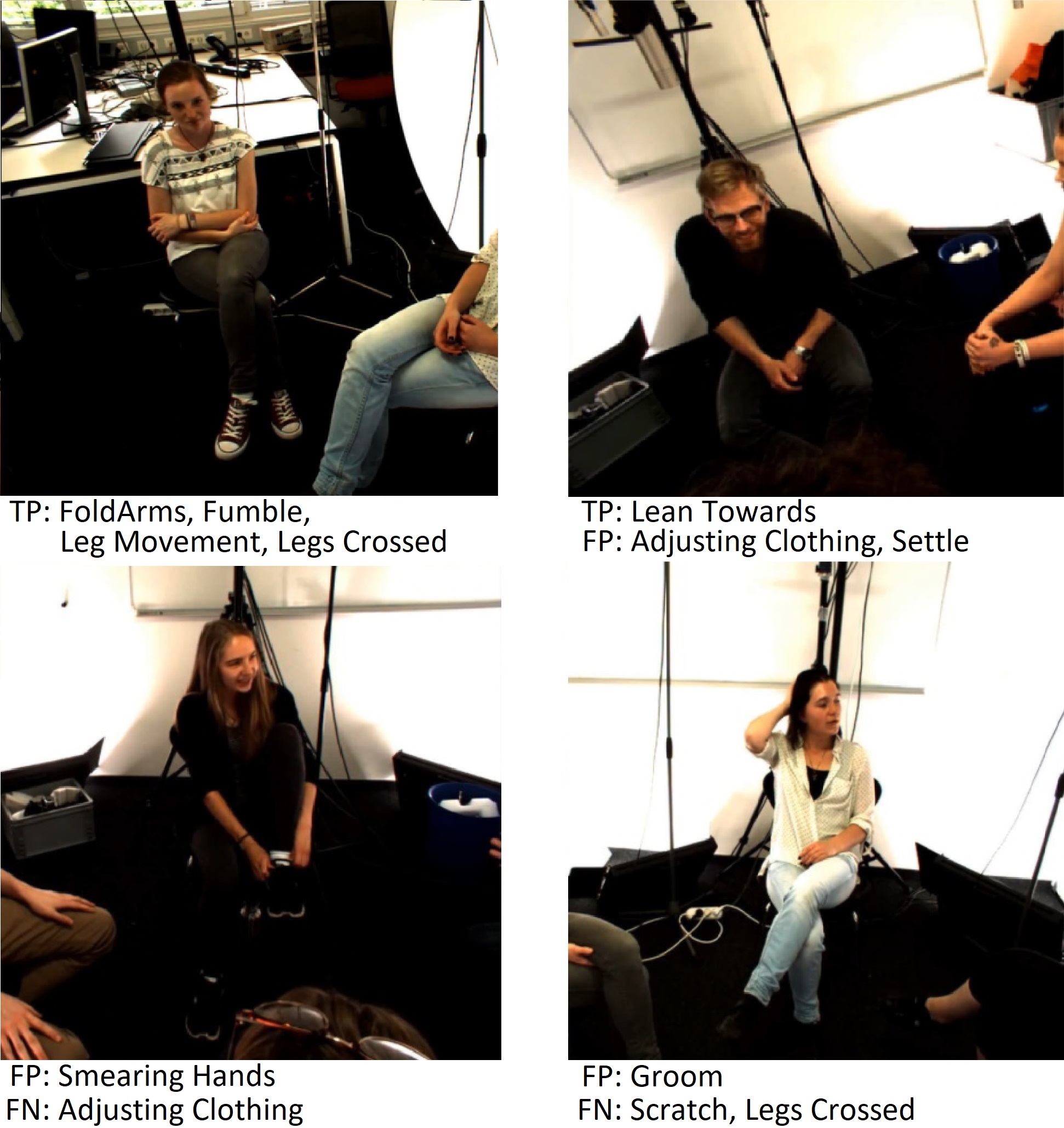}
    \vspace{-5pt} %%%
    \caption{Illustrative examples of true positive, false positive and false negative predictions.}
    \label{fig:examp}
    %\vspace{-5pt} %%%
\end{figure}

\subsection{Detection Evaluation}

In addition to MAP, the evaluation metrics are calculated from the detection confidence vector on the frame level: True Positives~(TP) as the sum of confidences in true classes, False Positives~(FP) as the sum of confidences in false classes, and False Negatives~(FN) as the sum of $1$ minus confidences in true classes.
%Dividing each of them by the number of summands, we obtain True Positive Rate~(TPR), False Positive Rate~(FPR) and False Negative Rate~(FNR), respectively, using micro and macro averaging.
From TP, TN and FN on the frame level, we calculate the F1 score globally using both micro and macro averaging.
As expected from behavior recognition, Swin achieves the best results with
%TPR 0.613~(micro) and 0.356~(macro) on Single-Label setting, and 0.568~(micro) and 0.321~(macro) on Multi-Label.
F1 0.728/0.544 and MAP 0.742/0.415 on the Single-Label setting, and F1 0.726/0.511 and MAP 0.691/0.367 on Multi-Label, using micro/macro averaging respectively.
Figure~\ref{fig:examp} illustrates examples of true positive, false positive and false negative predictions.
As in behavior detection without class balancing, there is a high inter-class performance variance. 
%As in behavior detection with no class balancing applied, there is a high inter-class variance between performance. 
The most frequent class \textit{Legs Crossed} reaches the highest performance among all classes.
%Namely the most frequent class \textit{Legs Crossed} reaches the highest performance, suppressing all other classes.

\begin{table*}[t]
    %\vspace{-5pt} %%%
    \caption{Evaluation of PDAN~\cite{dai2021pdan} with four types of features trained in both labeling settings in terms of F1 and MAP using micro/macro averaging.}
    \vspace{-10pt} %%%
    \begin{tabular}{l|cc|ccc|c}
    \toprule
    \multirow{2}{*}{Features} & \multicolumn{2}{c|}{Single-Label} & \multicolumn{2}{c}{Multi-Label} \\
    & F1 & MAP & F1 & MAP \\
    \midrule
    random & 0.348 / 0.114 & 0.312 / 0.092 & 0.348 / 0.115 & 0.312 / 0.092 \\
    I3D~\cite{8099985} & 0.542 / 0.340 & 0.502 / 0.276 & 0.581 / 0.339 & 0.557 / 0.271 \\
    TSN~\cite{TSN2016ECCV} & 0.550 / 0.309 & 0.624 / 0.291 & 0.531 / 0.343 & 0.658 / 0.311 \\
    TSM~\cite{lin2019tsm} & 0.660 / 0.419 & 0.600 / 0.311 & 0.627 / 0.347 & 0.626 / 0.293 \\
    Swin~\cite{liu2021Swin} & 0.729 / 0.545 & 0.742 / 0.415 & 0.726 / 0.511 & 0.691 / 0.367 \\
    \bottomrule
    \end{tabular}
    \label{tab:pdan}
    %\vspace{-5pt} %%%
\end{table*}

\section{Discussion}
\subsection{Annotations}

We presented the first publicly available annotations of 15 body language classes on a multi-view group discussion dataset.
Annotating human bodily behavior is challenging due to the subtle and often subjective nature of body language.
To evaluate the agreement of our annotators, we conducted a dedicated experiment.
%The overall annotation quality in terms of inter-annotator agreement on our dataset is comparable to numbers reported in previous work~\cite{liu2021imigue}.
Our class-based analysis of annotator agreement revealed clear differences between the agreement for different behavior classes, which should be taken into account by potential users of the dataset.
Restricting to a subset of the annotated classes to those with a proper relevance to the particular application and a high inter-annotator agreement can be a good practice for any body language analysis system.
On the other hand, if body language annotations are used to train a feature representation that is used in a downstream task, even low agreement classes can still be useful.
On the other hand, if body language annotations are used to train a feature representation that is used in a downstream task, even classes with low annotator agreement can still be useful.
We include classes with low agreement scores for full transparency and leave it to the users to decide which classes to use depending on their preferences. Supplementary material provides class-specific evaluation results to facilitate comparison with researchers who choose a subset of classes.

%As a result, we also include classes with low agreement scores for full transparency and leave it for the users of the dataset to decide which of the provided classes to use depending on their preferences with respect to reliability and coverage over behavior classes in their application. Supplementary material provides class-specific evaluation results to facilitate comparison with researchers who choose to use a reduced number of classes.

\subsection{Achieved Performance}

%Due to their performance, transformers are becoming more and more widespread in computer vision tasks~\cite{dosovitskiy2021an,khan2021}. Their effectiveness is also reflected in action recognition over our dataset. As even the Tiny version of Swin Transformers has outperformed all other CNN-based architectures in every setup where it was applied. This is usually followed by TSM and TSN, although I3D has a higher potential due to its 10-times larger number of parameters.

In line with the recent trend on computer vision tasks~\cite{dosovitskiy2021an,khan2021}, the effectiveness of transformers is also reflected on BBSI. Even the Tiny version of Swin Transformers has outperformed all other CNN-based architectures in every setup where it was applied. This is usually followed by TSM and TSN, although I3D has a higher potential due to its 10-times larger number of parameters. 

Class balancing degrades the performance in any setup. Although it was introduced to counteract the dominance of static classes, the MAP drop is the highest in those setups where the static classes are included. Our assumption is that equally balancing the dataset is not adequate in this case as the distribution of instance numbers per classes are exponential. Despite giving equal weight to the classes of very few instances increases their performance on the training set, they are not possible to achieve good recognition on unseen data. Not only it does not improve testing inference, the metrics of other classes fall as well.

Applying the Single-Label setting on a detection task inherently produces incorrect predictions. As in most of the cases there is at least one static class involved in concurrent actions, excluding static classes results in a classification problem of a significantly reduced rate of multiple labels. Thus, the difference between the Single-Label and Multi-Label experiments when the static classes are excluded is almost negligible compared to the case when all the classes are included, which is in the range between 0.005--0.088 MAP if there is no class balancing applied.

\subsection{Applications}

%Both our annotations as well as the presented prediction approaches create opportunities for a variety of applications.
The primary intended application for BBSI annotations is to train and evaluate algorithms that predict body language classes.
However, our annotations can also be useful in a pre-training step or for auxiliary training of approaches that address high-level behavior interpretation tasks such as leadership detection~\cite{beyan2017moving,muller2019emergent} or personality prediction~\cite{visapp22,palmero2021context} for which only limited amount of training data is available.
Furthermore, it can be of interest for behavioral scientists to use our annotations for research on the expression of nonverbal behavior in group interactions and how it relates to aspects like leadership, rapport, or interpersonal synchrony.

%In addition to our dataset, the prediction methods we evaluated can be applied in a variety of social signal processing and human-machine interaction scenarios.
%The predictions can serve as input features for social behavior interpretation tasks such as emotion recognition, emergent leadership detection, or personality prediction.
As BBSI is based on a rating scheme developed in the context of psychiatric interactions~\cite{troisi1999ethological}, we expect our body language predictions to be highly useful in clinical tasks, e.g. for depression detection~\cite{yin2019multi} or to estimate the quality of the therapist-patient relationship~\cite{horvath1993role}.
%In such scenarios, interpretability is an important property of any automatic method.
Using a set of psychologically motivated behaviors as an intermediate representation instead of generic pose-based features or deep learning representations will allow for better interpretability and build trust with clinicians and patients alike.
Our presented prediction methods can also be integrated into existing conversation analysis tools~\cite{penzkofer2021conan}, which at present do not have the ability to detect fine-grained body language.
%The prediction methods presented in this paper can also be integrated into existing conversation analysis tools~\cite{penzkofer2021conan}.
%At present, these tools do not have the ability to detect fine-grained classes of body language.

\subsection{Limitations and Future Work}

Our novel annotations and state-of-the-art evaluations represent an important step towards automatic analysis of body language in social interaction.
At the same time, several challenges remain that need to be addressed in future work.
While the BBSI set of behavior classes is motivated by previous work linking those classes to social attributes like leadership, rapport, or emotions, this link needs to be solidified by investigating the predictive power of bodily behaviors for such downstream tasks.
Furthermore, as the MPIIGroupInteraction dataset consists of participants recruited at a German university, future work should collect comparable datasets with more diverse cultural backgrounds.
%Furthermore, the MPIIGroupInteraction dataset consists of participants recruited at a German university, which restricts the datasets' cultural diversity.
%Future work should collect comparable datasets with participants of different cultural backgrounds.
A key challenge on BBSI is the large class imbalance that makes it difficult to train accurate models for classes that occur seldomly in natural behavior.
Future work could investigate generation of synthesized training examples or advanced data augmentation techniques.
%Future work could investigate the generation of synthesized training examples or on employing advanced data augmentation techniques.
The detection and classification approaches presented in this paper learn a single model that is applied to all participants.
%Concerning body language classification and detection, the approaches presented in this paper learn a single model that is applied to all participants.
While this is a meaningful first step to approach the task, the expression of body language is highly individual.
Future work should investigate personalization and test-time adaptation~\cite{amin2014test,charles2016personalizing} to model personal idiosyncrasies adequately.
%Future work should also investigate personalization and test-time adaptation~\cite{amin2014test,charles2016personalizing} to model personal idiosyncrasies adequately.
Another possible improvement is to use multi-channel inputs, e.g. by exploiting all three views on a person, or adding pose information~\cite{cao2021openpose}.
%Another possibility to improve future approaches is the use of multi-channel inputs, for example by exploiting all three views on a target person, or by using additional skeleton information from body pose estimation approaches~\cite{cao2021openpose}.

%\philipp{TODO - incorporate: Incorporating these predictions in approaches for different downstream tasks is beyond the scope of this paper and left to future work.}

%\philipp{Todo for all: put additional ideas for limitations and future work here} \michal{Limitation: Class unbalance. Future work: Prediction models for leadership/rapport/liking,...? Potential for using multi-channel inputs such as skeleton, transcript, audio?}

\section{Conclusion}
In this work, we presented BBSI, the first publicly available set of annotations of subtle bodily behaviors in group interactions.
The novel annotations consist of 15 body language classes that were densely annotated for 26 hours of human behavior recorded from 78 participants on the publicly available MPIIGroupInteraction dataset.
We provided results of descriptive analyses of the annotations as well as a dedicated experiment on annotation quality, as they were done manually by our human annotators.
Furthermore, we presented the results of state-of-the-art action recognition approaches evaluated on the MPIIGroupInteraction dataset with the BBSI annotations.
As such, our work is a key contribution to advance in-depth analyses of subtle body language cues in human interactions.

\begin{acks}
This work was supported by the \grantsponsor{}{French National Research Agency}{https://anr.fr/} under the UCA\textsuperscript{JEDI} Investments into the Future, project number \grantnum{}{ANR-15-IDEX-01}, and by the \grantsponsor{}{German Ministry for Education and Research}{https://www.bmbf.de/}, grant number \grantnum{}{01IS20075}.
\end{acks}

\balance % required
\bibliographystyle{ACM-Reference-Format}
\bibliography{egbib}

\end{document}

\title[Bodily Behaviors in Social Interaction: Novel Annotations and State-of-the-Art Evaluation]{Bodily Behaviors in Social Interaction:\\Novel Annotations and State-of-the-Art Evaluation}

\subtitle{Supplementary Material}

\author{Michal Balazia}
\authornote{Authors contributed equally.}
\email{michal.balazia@inria.fr}
\orcid{0000-0001-7153-9984}
\affiliation{%Méditerranée
  \institution{INRIA Sophia Antipolis}
  \streetaddress{2004 Route des Lucioles}
  \city{Sophia Antipolis}
  \country{France}
  \postcode{06902}
}
\author{Philipp Müller}
\authornotemark[1]
\email{philipp.mueller@dfki.de}
\orcid{0000-0001-7037-7100}
\affiliation{%
  \institution{DFKI Saarbrücken}
  \streetaddress{Stuhlsatzenhausweg 3}
  \city{Saarbrücken}
  \country{Germany}
  \postcode{66123}
}
\author{Ákos Levente Tánczos}
\email{akos.tanczos@inria.fr}
\orcid{0000-0002-4760-9599}
\affiliation{%
  \institution{INRIA Sophia Antipolis}
  \streetaddress{2004 Route des Lucioles}
  \city{Sophia Antipolis}
  \country{France}
  \postcode{06902}
}
\author{August von Liechtenstein}
\email{august.liechtenstein@dfki.de}
\orcid{0000-0002-8774-746X}
\affiliation{%
  \institution{DFKI Saarbrücken}
  \streetaddress{Stuhlsatzenhausweg 3}
  \city{Saarbrücken}
  \country{Germany}
  \postcode{66123}
}
\author{François Brémond}
\email{francois.bremond@inria.fr}
\orcid{0000-0003-2988-2142}
\affiliation{%
  \institution{INRIA Sophia Antipolis}
  \streetaddress{2004 Route des Lucioles}
  \city{Sophia Antipolis}
  \country{France}
  \postcode{06902}
}
\renewcommand{\shortauthors}{Balazia and Müller, et al.}

%% http://dl.acm.org/ccs.cfm
\begin{CCSXML}
<ccs2012>
<concept>
<concept_id>10010147.10010178.10010224.10010226</concept_id>
<concept_desc>Computing methodologies~Image and video acquisition</concept_desc>
<concept_significance>500</concept_significance>
</concept>
<concept>
<concept_id>10010147.10010178.10010224.10010225.10010228</concept_id>
<concept_desc>Computing methodologies~Activity recognition and understanding</concept_desc>
<concept_significance>500</concept_significance>
</concept>
<concept>
<concept_id>10003120.10003130</concept_id>
<concept_desc>Human-centered computing~Collaborative and social computing</concept_desc>
<concept_significance>300</concept_significance>
</concept>
<concept>
<concept_id>10010405.10010455.10010459</concept_id>
<concept_desc>Applied computing~Psychology</concept_desc>
<concept_significance>300</concept_significance>
</concept>
</ccs2012>
\end{CCSXML}
\ccsdesc[500]{Computing methodologies~Image and video acquisition}
\ccsdesc[500]{Computing methodologies~Activity recognition and understanding}
\ccsdesc[300]{Human-centered computing~Collaborative and social computing}
\ccsdesc[300]{Applied computing~Psychology}

\keywords{dataset, body pose, gesture, social signals, behavior detection}

\maketitle

%\section{Connection to High-Level Attributes}
%We investigate the connections to 
%Figure~\ref{fig:bl_to_high_level}
%
%\begin{figure*}[h]
%    \includegraphics[width=0.49\textwidth]{fig/Duration_PLead.pdf}
%    \includegraphics[width=0.49\textwidth]{fig/Duration_rapport.pdf}
%    \includegraphics[width=0.49\textwidth]{fig/Duration_PLead.pdf}
%    \includegraphics[width=0.49\textwidth]{fig/Duration_rapport.pdf}
%    \caption{Correlations between the the frequency of body language classes shown by participants and received leadership (left) and rapport ratings (right). The dotted line indicates significance of the correlation coefficient at $\alpha = 0.05$ (not corrected for multiple tests). \philipp{add -0.5 on y axis} \michal{or -0.3 if it looks better. You can also substitute the 3rd and 4th plot with liking and turn-taking.}}
%    \label{fig:bl_to_high_level}
%\end{figure*}

\section{Confidence Matrices of Behavior Recognition Methods}

\begin{figure*}[h]
    \subfigure[Single-Label I3D~\cite{8099985}]{\includegraphics[width=0.475\textwidth]{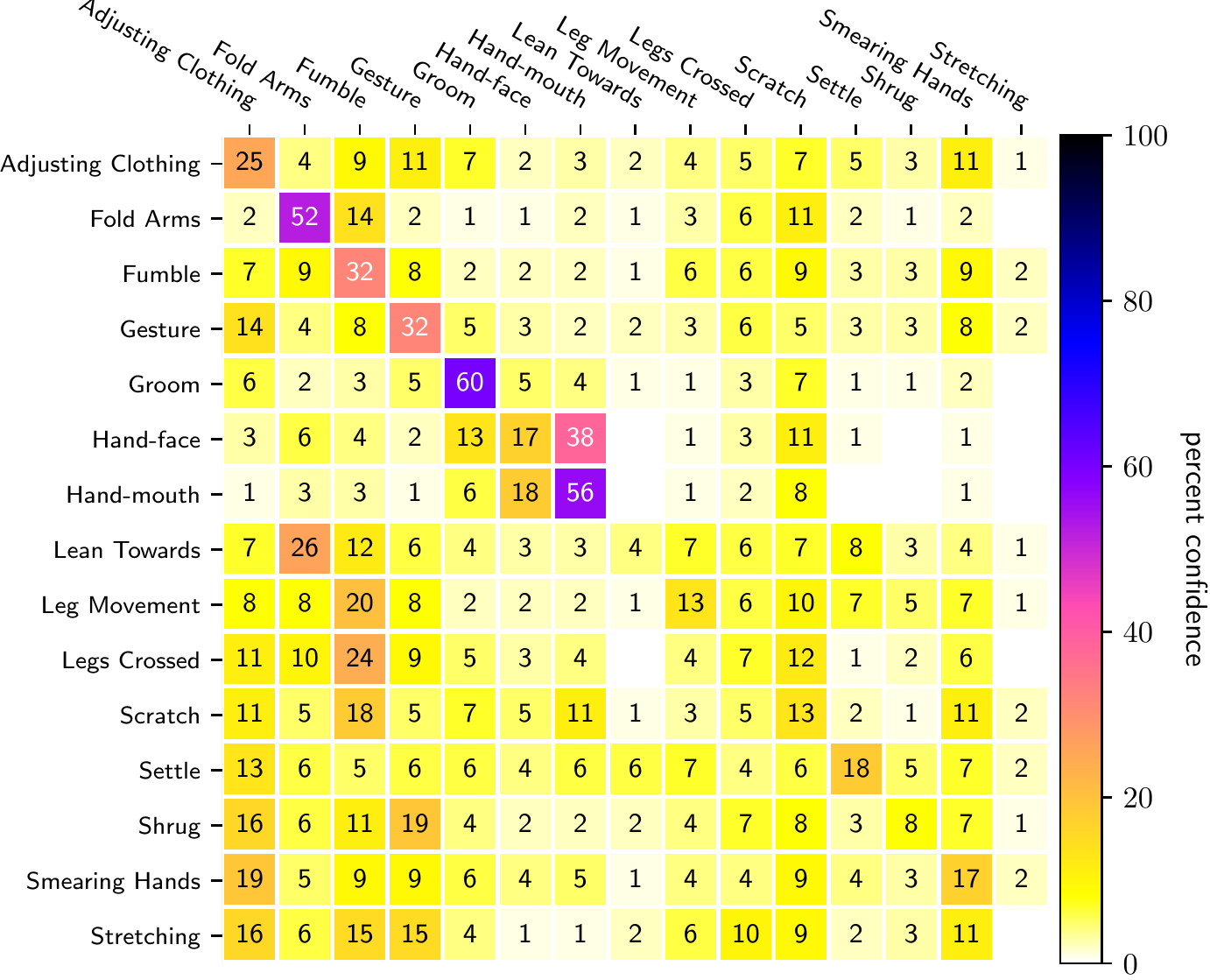}}
    \hspace{20pt}
    \subfigure[Multi-Label I3D~\cite{8099985}]{\includegraphics[width=0.475\textwidth]{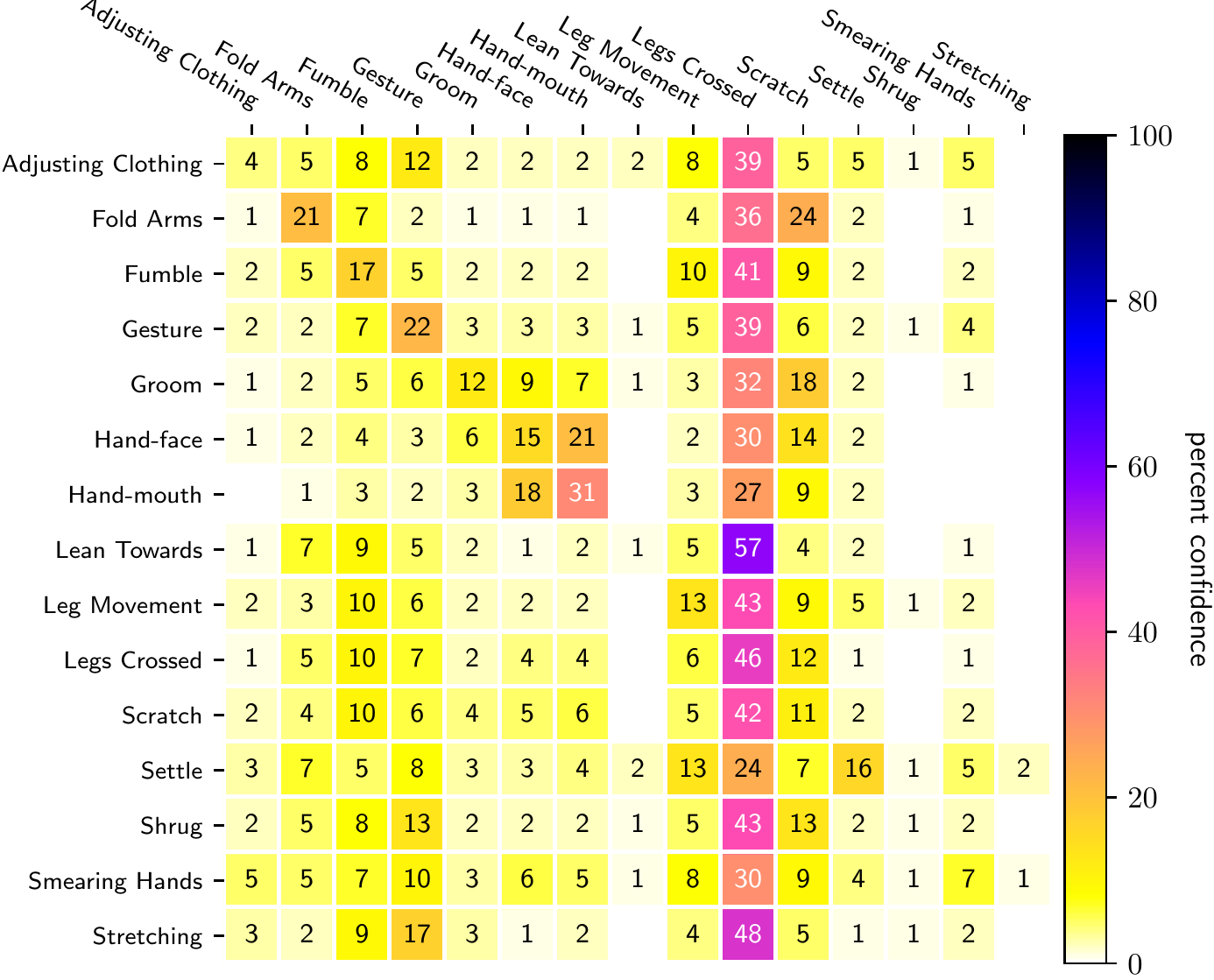}}
    \subfigure[Single-Label TSN~\cite{TSN2016ECCV}]{\includegraphics[width=0.475\textwidth]{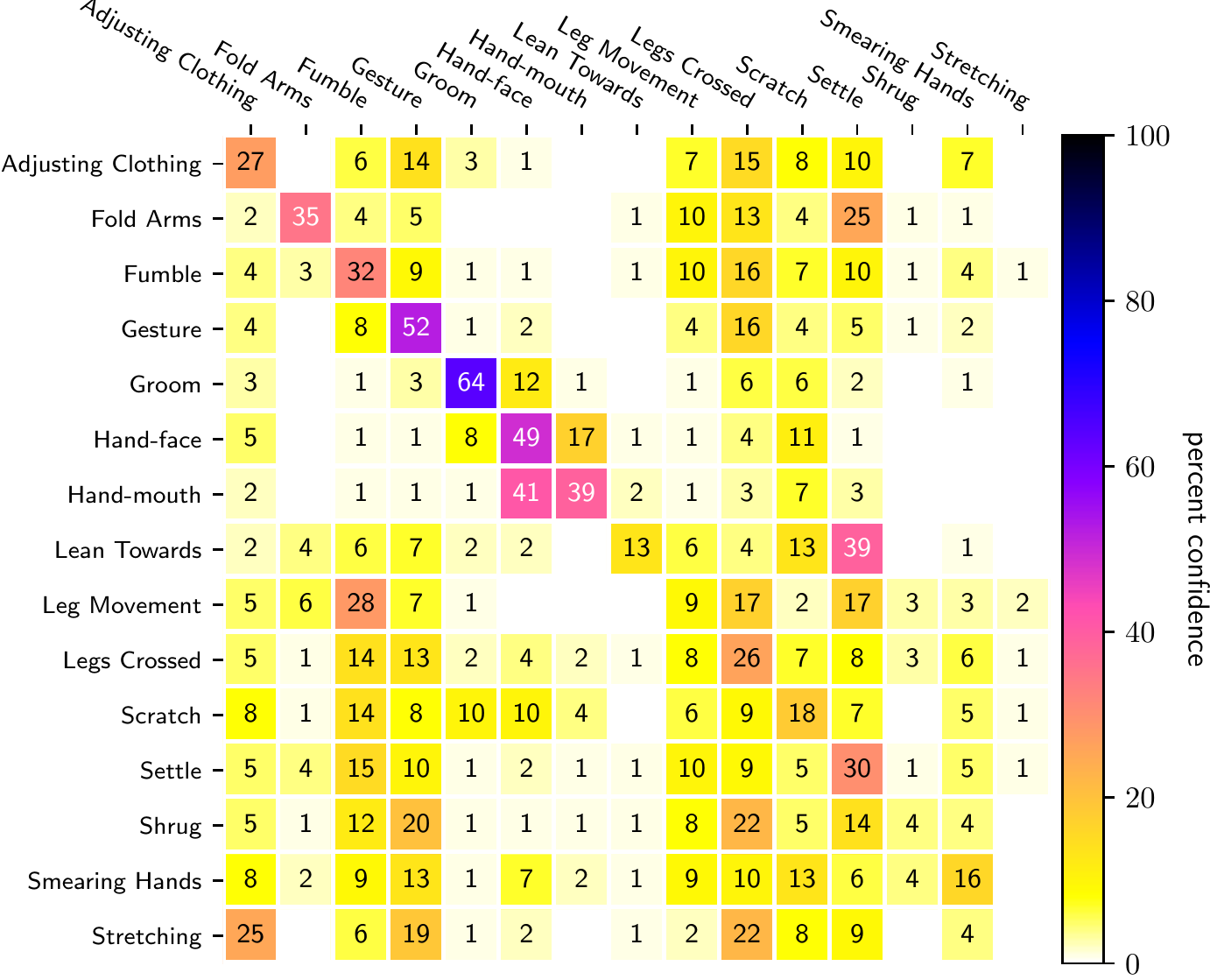}}
    \hspace{20pt}
    \subfigure[Multi-Label TSN~\cite{TSN2016ECCV}]{\includegraphics[width=0.475\textwidth]{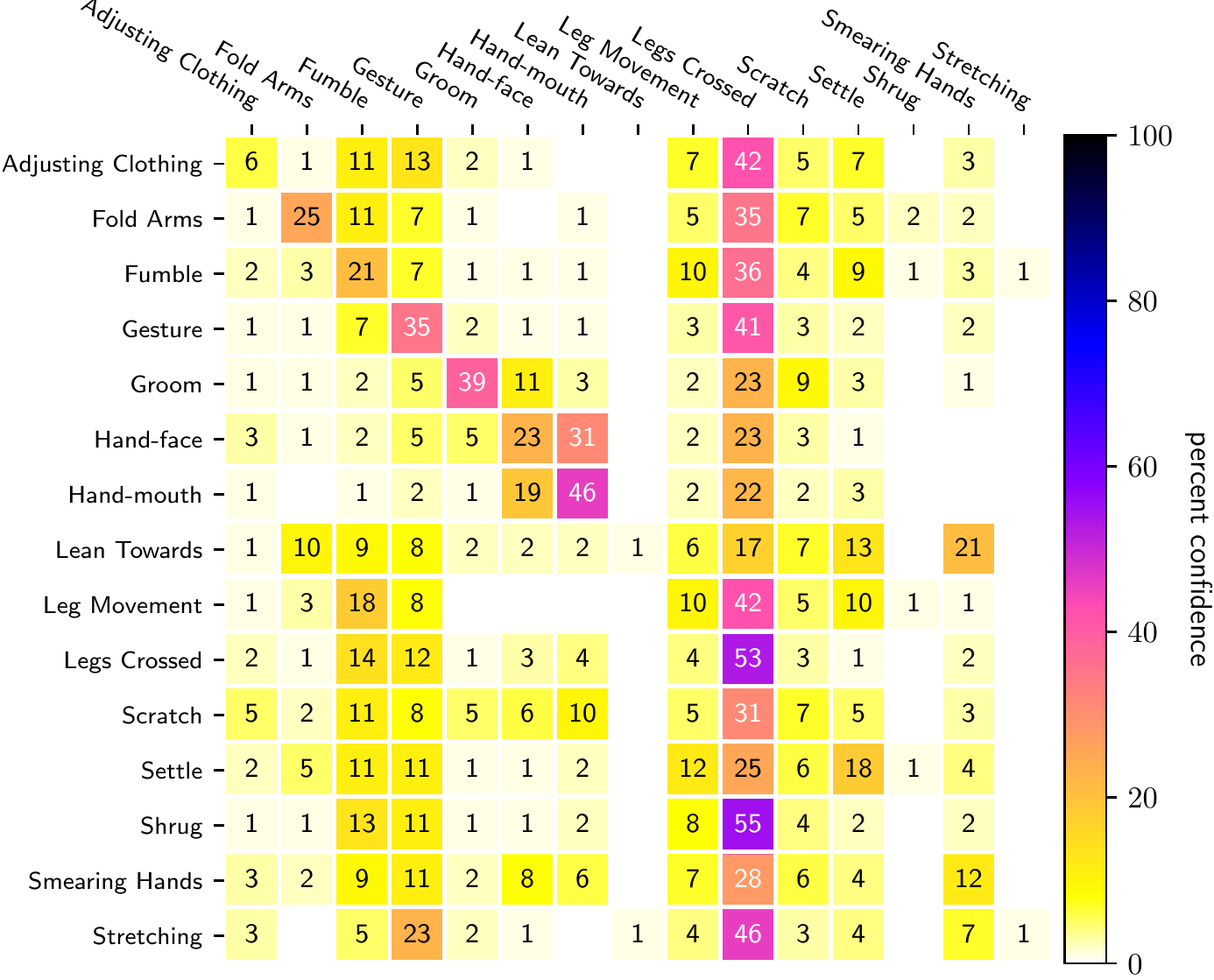}}
    \caption{Confidence matrices of behavior recognition by I3D and TSN, trained in both labeling settings.}
    \label{fig:conf1}
\end{figure*}

\begin{figure*}[h]
    \vspace{-5pt}
    \subfigure[Single-Label TSM~\cite{lin2019tsm}]{\includegraphics[width=0.475\textwidth]{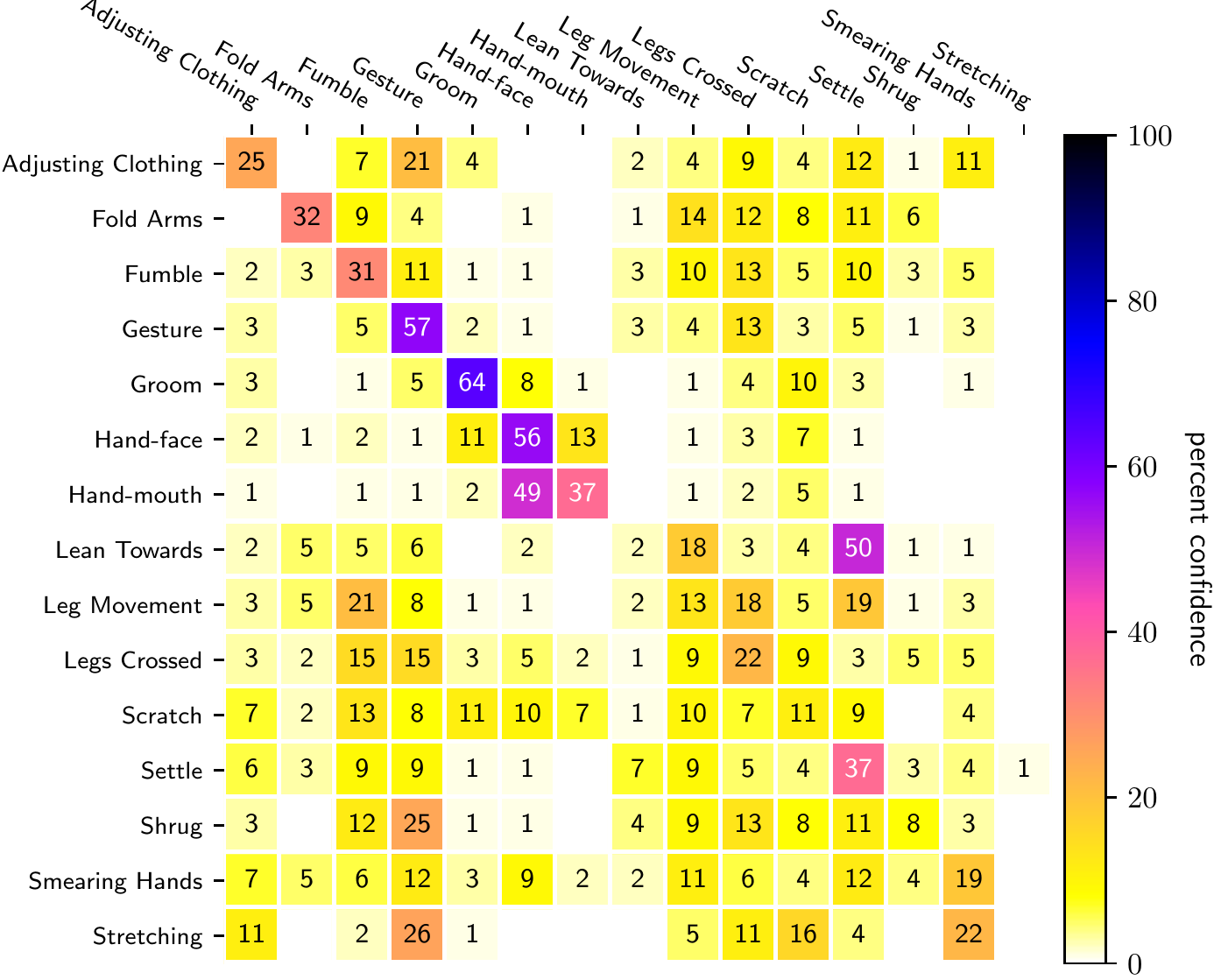}}
    \hspace{20pt}
    \subfigure[Multi-Label TSM~\cite{lin2019tsm}]{\includegraphics[width=0.475\textwidth]{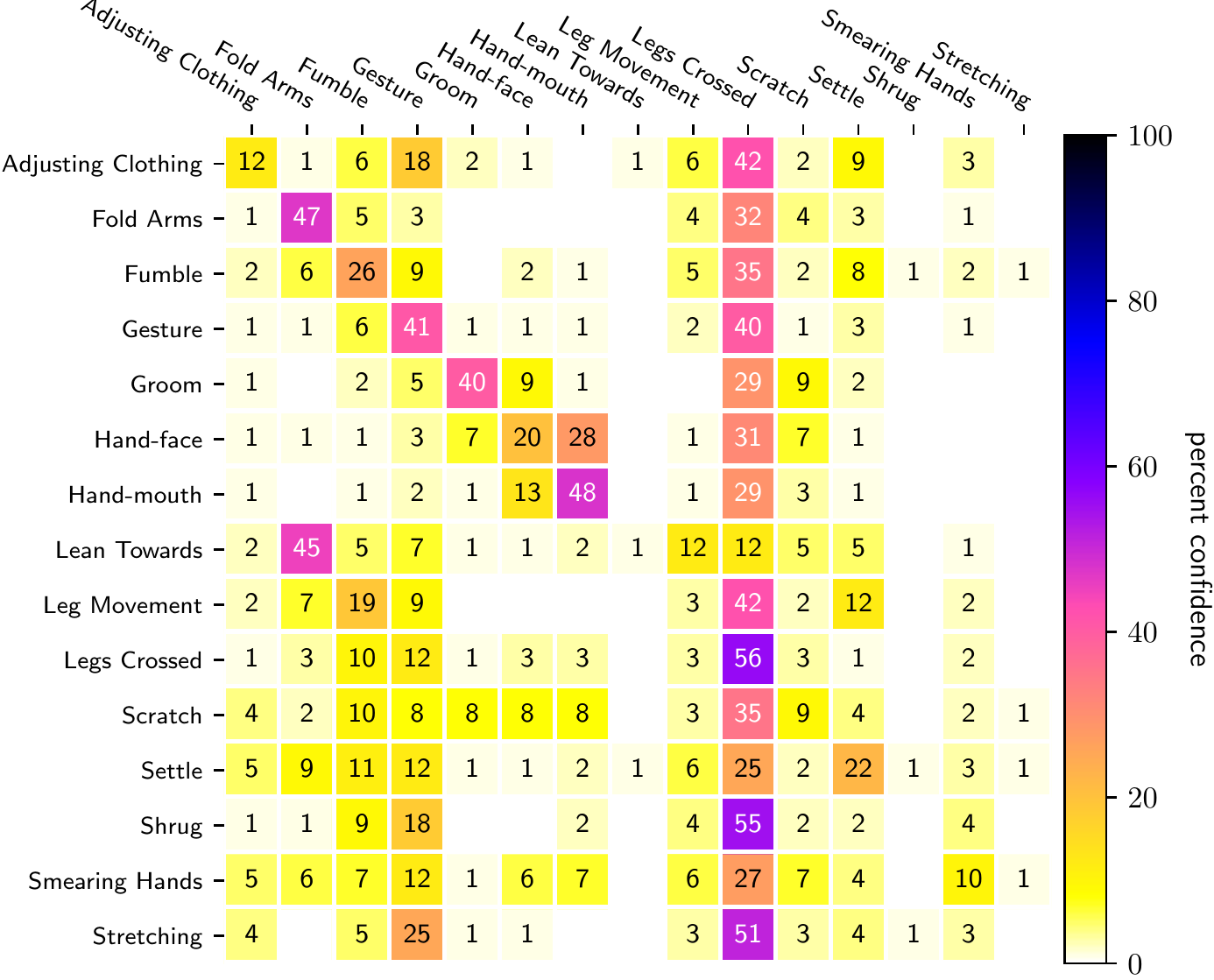}}
    \subfigure[Single-Label Swin~\cite{liu2021Swin} ]{\includegraphics[width=0.475\textwidth]{fig/matrix-sin-swin.pdf}}
    \hspace{20pt}
    \subfigure[Multi-Label Swin~\cite{liu2021Swin} ]{\includegraphics[width=0.475\textwidth]{fig/matrix-mul-swin.pdf}}
    \caption{Confidence matrices of behavior recognition by TSM and Swin, trained in both labeling settings.}
    \label{fig:conf2}
\end{figure*}

Evaluation of all methods can be visualized by aggregating all confidence vectors into a confidence matrix. See Figure~\ref{fig:conf1} and Figure~\ref{fig:conf2} for the confidence matrices of all four behavior recognition algorithms.

\section{Illustrative Examples of PDAN Predictions}

\begin{figure*}[ht]
    \includegraphics[width=1\textwidth]{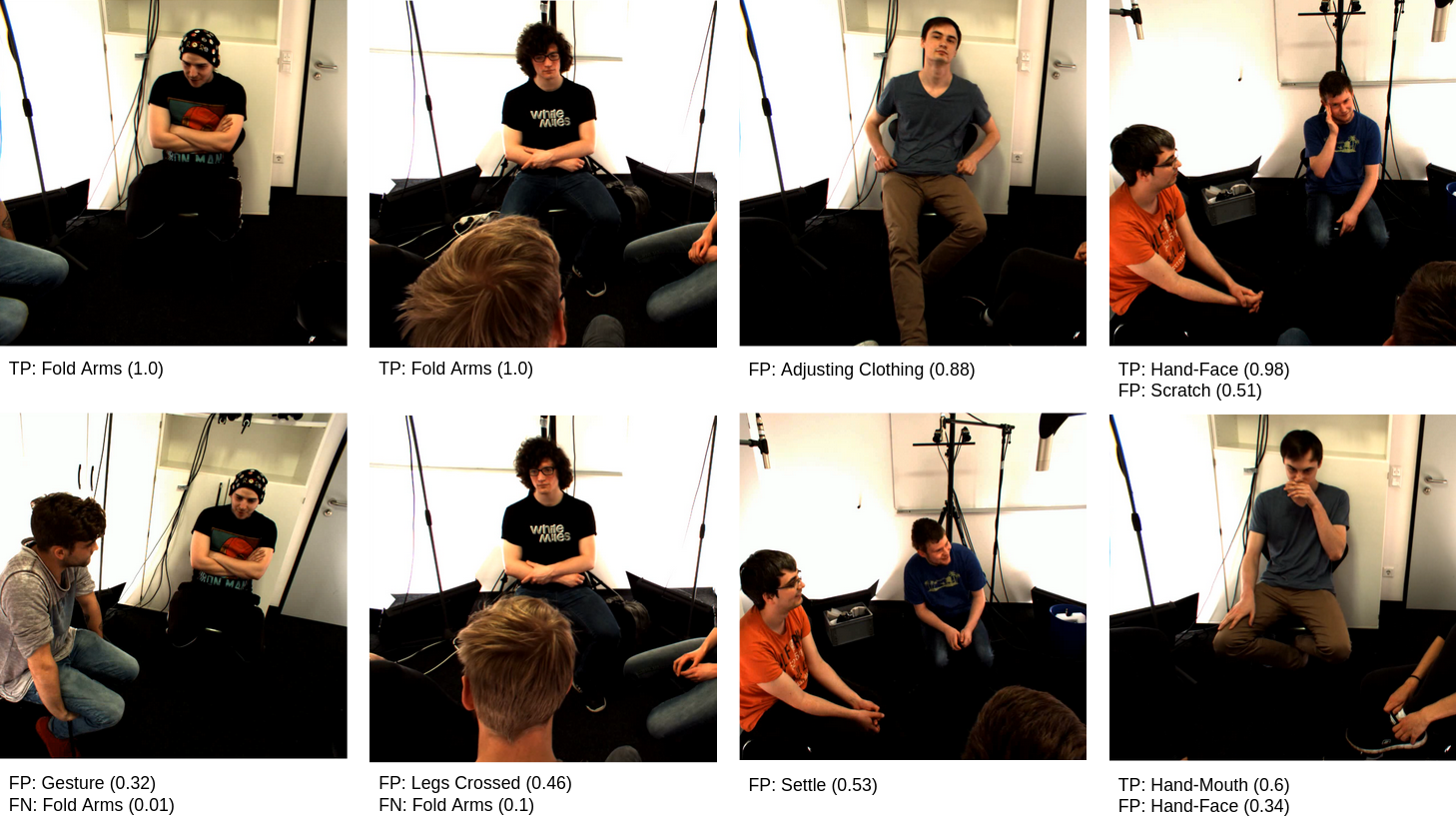}
    \caption{Illustrative examples of PDAN predictions on the frame level. Ground truth and acceptance threshold of $0.3$ determine one of the three statuses: True Positive~(TP), False Positive~(FN) and False Negative~(FN). First column shows a decrease in performance caused by an additional person present in the scene. Second column shows the effect of occlusion. Third column indicates that classes with lower level of agreement have a generally worse performance. Fourth column highlights the great challenge of distinguishing between classes of very subtle differences.}
    \label{fig:conf}
\end{figure*}

We analyzed four major sources of impact on performance: additional person in the scene, occlusion, low class agreement, distinction between visually similar classes. Figure~\ref{fig:conf} illustrates these examples in the context of ground truth and prediction.

\bibliographystyle{ACM-Reference-Format}
\bibliography{egbib}